\documentclass[ijds,nonblindrev]{informs-ijds}

\OneAndAHalfSpacedXI

\usepackage{natbib}
 \bibpunct[, ]{(}{)}{,}{a}{}{,}%
 \def\bibfont{\small}%
\TheoremsNumberedThrough     
\ECRepeatTheorems

\EquationsNumberedThrough    

\MANUSCRIPTNO{IJDS-0001-1922.65}

\usepackage{algorithmic}
\usepackage{algorithm}
\usepackage{booktabs}
\usepackage{adjustbox}

\begin{document}


\RUNAUTHOR{Kosolwattana et al.} 

\RUNTITLE{Online Modeling and Monitoring for Dependent Dynamic Processes under Resource Constraints}

\TITLE{Online Modeling and Monitoring for Dependent Dynamic Processes under Resource Constraints}

\ARTICLEAUTHORS{%
\AUTHOR{Tanapol Kosolwattana}
\AFF{Department of Industrial Engineering, University of Houston, \EMAIL{tkosolwattana@uh.edu}}
\AUTHOR{Huazheng Wang}
\AFF{School of Electrical Engineering and Computer Science, Oregon State University, \EMAIL{huazheng.wang@oregonstate.edu}}
\AUTHOR{Ying Lin}
\AFF{Department of Industrial Engineering, University of Houston, \EMAIL{ylin53@central.uh.edu}}
} 

\ABSTRACT{%
Adaptive monitoring of a large population of dynamic processes is critical for the timely detection of abnormal events under limited resources in many healthcare and engineering systems. Examples include the risk-based disease screening and condition-based process monitoring. However, existing adaptive monitoring models either ignore the dependency among processes or overlook the uncertainty in process modeling. To design an optimal monitoring strategy that accurately monitors the processes with poor health conditions and actively collects information for uncertainty reduction, a novel online collaborative learning method is proposed in this study. The proposed method designs a collaborative learning-based upper confidence bound (CL-UCB) algorithm to optimally balance the exploitation and exploration of dependent processes under limited resources. Efficiency of the proposed method is demonstrated through theoretical analysis, simulation studies and an empirical study of adaptive cognitive monitoring in Alzheimer's disease. 
}%


\maketitle
%
\section{Introduction}
We are investigating a large-scale population consisting of $N$ units. Each of these units exhibits a distinct and dynamic health progression process. Due to limited monitoring or sensing resources, only $M$ (where $M < N$) out of the $N$ units can be monitored in each monitoring cycle. Our interest lies in allocating the monitoring resources to the high-risk units with severe health conditions to ensure timely initiation of preventive care \citep{8169076, gartner2015machine}. This problem is pervasive across multiple sectors, encompassing domains such as healthcare \citep{bai2022surgery, gartner2015machine, lowe2004monitoring} and manufacturing \citep{abbou2019group, liang2013opportunistic}. In the context of cognitive monitoring of Alzheimer's Disease (AD), for instance, each unit corresponds to a patient. Prioritizing the monitoring of patients experiencing rapid cognitive decline or impairment can offer valuable insights for instigating disease treatment in its nascent stages, consequently mitigating potential cognitive deterioration \citep{lazarou2016novel, weimer2009early}. Resource constraints, such as limited staff availability and appointment slots within the healthcare system, underscore the necessity of directing restricted monitoring resources towards patients with the most severe cognitive conditions \citep{sharif2023priority, doi:10.1080/24725854.2018.1470357}. In battery systems, in order to understand the health state accurately, skilled professionals need to remove batteries from the electrical vehicles and measure the internal resistance level and the full charge capacity \citep{friansa2017development, rahimi2013battery}. It is operationally costly to exhaustively inspect a large number of batteries on a regular basis \citep{mysorewala2022review}. However, accurately allocating monitoring resources to severe units remains a challenging issue. This is primarily due to the dynamic evolution of their health status over time. The progression dynamics are unknown in prior and hard to be estimated from insufficient monitoring data, especially at the beginning of monitoring. Therefore, it is in urgent need to develop novel online modeling and monitoring method that real-time estimate the process dynamics and adaptively allocate monitoring resources to severe ones. 

Allocating limited resources within uncertain environments has frequently been formulated as a class of  multi-armed bandit (MAB) problems in the existing literature \citep{lattimore2015linear, ahuja2020approximation}. The fundamental objective of MAB problems is to distribute constrained resources among a subset of options, often referred to as "arms", in a manner that maximizes the expected rewards during each decision cycle. Given that the reward associated with each arm is not known beforehand, MAB algorithms—such as the linear upper confidence bound (LinUCB)\citep{li2010contextual}—determine the optimal allocation strategy by carefully balancing the interplay between exploitation (leveraging arms with the highest predicted rewards) and exploration (gaining more information to enhance the accuracy of model predictions) \citep{han2020differentially, 10.1145/2911451.2911528}. However, the application of MAB algorithms to the realm of online modeling and monitoring of dynamic processes introduces several intricate challenges.

First, conventional MAB algorithms were developed for static arms whose underlying reward distributions remain unchanged over time \citep{besbes2014stochastic, garivier2011upper, auer2002finite}. These algorithms were primarily designed to estimate the reward distributions and identify the arm with the highest expected reward. In website advertising placement, for instance, the primary objective is to maximize profits by selecting the commercial advertisement with the highest reward. This reward is typically measured by click-through rate (CTR) \citep{li2010contextual}, which is determined by a binary reward system: it is 1 when the advertisement is clicked and 0 otherwise. In cases where an advertisement campaign has been running for an extended period, the CTR can be treated as a static arm. This is because its reward distribution remains relatively stable, with no significant fluctuations or seasonality. However, in the case of health monitoring, the reward of monitoring is associated with the health condition of a process, which dynamically changes with respect to time or time-dependent predictors \citep{8169076}. These conventional MAB algorithms are inadequate to predict the monitoring reward from dynamic health progression and gather processes' information to infer the health progression dynamics.

Recently, there are some MAB algorithms that can deal with dynamic environment. \citep{dzhoha2023multi,kamienny2020learning}. 
For example, the dynamic environment in \citep{kamienny2020learning} is defined as the tasks for selection change over time. However, it does not match with our problem since our problem focuses on monitoring the same patient population. Also, these MAB algorithms were developed for a population of independent arms, making them challenging to be extended for a population of dependent processes \citep{wang2018online, 10.1145/2911451.2911528}. The dependency between dynamic processes is commonly observed in various healthcare and engineering systems \citep{8363633}. In the cognitive monitoring of the aging population, for example, the mechanisms of cognitive decline can be approximately classified into three types: normal aging, mild cognitive impairment, and Alzheimer's disease \citep{petersen2010alzheimer, petersen1999mild}. Patients with the same mechanism are likely to exhibit similar cognitive decline processes, whereas patients belonging to different types showcase distinct cognitive degradation patterns. On the other hand, cognitive degradation is affected by a set of risk factors, such as age, gender, and genes \citep{gyanwali2019risk, 8169076}. Patients with similar profiles on these risk factors also tend to exhibit similar cognitive degradation dynamics. Numerous studies have demonstrated the criticality of explicitly capturing the dependency between dynamic processes to enhance the accuracy of health progression modeling, particularly when dealing with limited training data \citep{doi:10.1080/24725854.2018.1470357, kumar2013flexible}. These studies often represented the dependency between processes as latent group structures at the population-level or the pairwise similarities between processes. However, existing methods that leverage the dependency between processes primarily focus on retrospective health progression modeling approaches, such as latent growth modeling \citep{peterson2011psychological, meng2022fine}, mixture models \citep{bradley1999mathematical, wang2015multivariate}, and the recently developed collaborative learning method \citep{8169076}. The development of algorithms for online modeling and monitoring of dependent processes is currently insufficient.
  
To mitigate these gaps, we propose an Online Collaborative Learning (OCL) algorithm that adaptively allocates the monitoring resources to a population of dependent processes through online modeling and monitoring. It compromises of two key phases: the health progression modeling of dependent processes from sequentially collected data and the sequential monitoring resources allocation under uncertainty and dependency. Initially, the health progression modeling phase is focused on capturing the inherent group structure within health progression dynamics and the interdependency among various units. This is achieved through a collaborative learning framework, allowing for an explicit representation of these relationships. The second phase introduces a groundbreaking Collaborative Learning-based upper confidence bound (CL-UCB) algorithm for adaptive monitoring resources allocation. This algorithm serves to develop a novel exploration strategy for dependent and dynamic processes, grounded in the concept of upper confidence bounds. The CL-UCB algorithm is instrumental in devising an optimal monitoring policy, which strikes a balance between exploiting high-risk units characterized by severe health conditions and exploring the uncertain nature of health progression dynamics. In summary, our proposed Online Collaborative Learning (OCL) algorithm bridges the identified gaps by offering a comprehensive framework for adaptively allocating monitoring resources to dependent processes. This is achieved through an integrated approach that encompasses health progression modeling and resource allocation while accommodating uncertainty and dependency.

The contributions of this paper can be summarized as:
\begin{itemize}
    \item We propose a novel Online Collaborative Learning (OCL) framework to explicitly capture the inherent group structure and similarities in health progression of dependent processes and harness the dependencies to design optimal monitoring strategy which balances the exploitation of high-risk units and the exploration of uncertain units via a new Collaborative Learning-based UCB (CL-UCB) algorithm.
    \item Through rigorous theoretical analysis, we demonstrate that accounting for process dependency leads to an improved upper regret bound for the monitoring strategy compared to alternative Multi-Armed Bandit (MAB) algorithms..
    \item We demonstrate the effectiveness of the proposed method through simulation studies and an empirical study of adaptive cognitive monitoring in Alzheimer's disease. The results showcase the improvements in parameter estimation and monitoring efficacy compared to alternative MAB algorithms.
\end{itemize}

The structure of this paper is organized as follows. Section II reviews the related works for this study. Section III introduces the proposed online collaborative learning (OCL) framework and the theoretical analysis of the proposed method. In Section IV, the efficiency of the proposed OCL method is demonstrated through a simulation study and an empirical study of cognitive degradation monitoring for Alzheimer's disease (AD). Finally, the conclusion of this paper is drawn in Section V.

\section{Literature Review}
Section \ref{sub2.1} reviews the retrospectively health progression modeling for dependent processes. Section \ref{sub2.2} summarizes the existing adaptive monitoring algorithms most of which were developed at the individual-level. Lastly, the review of existing multi-armed bandit algorithms is provided in Section \ref{sub2.3}.

\subsection{Retrospective Health Progression Modeling}
\label{sub2.1}
Modeling the health progression for a population of dependent units has been widely studied retrospectively using fixed and complete training data sets that cover the whole history of process progression \citep{yu2022healthnet, doi:10.1080/24725854.2018.1470357}. The first type of methods relies on the latent cluster structure assumption, which clusters the units to a set of latent progression trajectories based on the similarities in their health progression dynamics. To explicitly capture the latent cluster structure in health progression dynamics, the finite mixture models \citep{gerdtham2001equity}, latent class regression model \citep{streur2018atrial, vermunt2002latent}, and clusterwise linear regression \citep{chen2022cluster, zamaninasab2022cluster, desarbo1988maximum} have been developed. However, these statistical models usually rely on the distribution assumptions of latent variables, which can lead to misspecification issues on real data.
Moreover, they have high computational cost in model estimation, as they often employ the Expectation-Maximization (EM) algorithm which is sensitive to a high dimensional data and model complexity. These limitations may make the models hard to be used for online modeling and monitoring. To relax the distribution assumptions, a collaborative regression model was recently developed by formulating the health progression modeling of dependent units as a constrained matrix factorization problem, which can be efficiently solved by the gradient-based algorithm \citep{8169076}. It is also flexible to incorporate prior knowledge, such as the similarities between units, to enhance the estimation of model parameters. 
However, the collaborative regression model was still developed for retrospective modeling that require sufficient historical data to understand the whole process of health progression and cannot be used for online health modeling and monitoring. In this paper, we propose a novel online collaborative learning method to estimate the health progression dynamics of dependent units from sequentially collected data.

\subsection{Adaptive Monitoring}
\label{sub2.2}
An adaptive monitoring algorithm is a sequential decision making method that adjusts its monitoring strategies in real-time to optimize the resource allocation. The partially observable Markov decision process (POMDP) or Markov decision process (MDP) models are widely used techniques to solve adaptive monitoring or screening problems \citep{zang2020cmdp}. For example, a POMDP model was developed in \citep{10.2307/23323677} to inform personalized screening strategies of breast cancer to maximize the total expected quality-adjusted life years of patients. Another POMDP model in \citep{papakonstantinou2014planning} was introduced to create concrete corrosion deterioration inspecting strategies to minimize structural life-cycle costs. However, these methods usually assume the health progression dynamics are well known or only focus on the decision making of single unit, which are hard to be used for designing resource allocation strategies in a population of competing units. 
The selective sensing method developed in \citep{doi:10.1080/24725854.2018.1470357} is the only approach that monitors a population of dependent and uncertain units by integrating the health progression modeling part to predict units' health in real time and the sensing part to allocate resources to units that have the most severe health conditions. Even though this approach considers the dependencies between processes and updates the estimated health progression dynamics with newly collected information, it did not consider the exploration of uncertain environment to achieve the optimal exploitation-exploration trade-off in resource allocation. In this paper, we design an upper confidence bound (UCB)-based exploration strategy to optimally allocate the limited monitoring resources for balancing the exploitation and exploration of dependent dynamic processes. 

\subsection{Multi-armed bandit algorithms}
\label{sub2.3}
Multi-armed bandit algorithms provide the framework for decision-making over time under uncertainty. However, existing bandit algorithms cannot solve the adaptive monitoring problem defined in this paper because they either ignore the dynamic nature of units or do not exploit the latent structure in monitoring units. Contextual bandit algorithms were commonly used to address the explore-exploit dilemma when allocating the resources to high-reward units. The LinUCB algorithm determines the expected reward via the content features extracted from units \citep{10.1145/2911451.2911528}. However, because it estimates the unknown bandit parameters for each unit independently, it ignores dependency among units, meaning it does not exploit the shared information between units and creates bias during the learning process. To tackle this problem, the GOB.Lin algorithm introduced in \citep{DBLP:journals/corr/Cesa-BianchiGZ13} connects units as a network through a graph Laplacian regularization which shares the contexts and rewards among units. It assumes that the parameters of units related to each other are close. However, the unknown parameters estimated by the GOB.Lin algorithm do not represent a complex heterogeneous structure which is common in various health progression systems. Apart from GOB.Lin, the Collaborative Linear Bandit or CoLin in short developed in \citep{10.1145/2911451.2911528} also integrates similarity information from users to the model. While GOB.Lin requires connected units in the network to have similar bandit parameters through graph Laplacian regularization, CoLin instead makes the assumptions about the reward generation via an additive model. Specifically, it allows neighboring units to share their influence on neighbors' decisions which provide the capability to capture the heterogeneity among units.

There are a few MAB algorithms developed for dependent arms based on the concept of matrix factorization with low-rank matrix completion. To check if the arms are related to each other, the MLinGreedy in \citep{yang2021impact} assumes there is a shared linear feature extractor that maps the contextual features observed on different arms to a low-dimensional embedding. In the Hidden Linear Bandit algorithm (hLinUCB) developed in \citep{10.1145/2983323.2983847}, in addition to the observed contextual features, it learns the hidden features which are introduced to reduce bias in the reward generation since some of them are not detectable by the algorithm. Also, it applies the coordinate descent algorithm to estimate the hidden features and the unknown bandit parameters and to derive the exploration strategy for online learning. Although our algorithm also leverages the matrix factorization approach to represent how similar each process is to the other, it is fundamental different with these existing MAB algorithms since 1) it deals with dependent dynamic processes by capturing the latent group structure in model parameters instead of observed features; 2) it further considers similarity regularization to model dependency among the dynamic processes; 3) it does not rely on the strict assumption of the assignment of minimum eigenvalue. Thus, it can be applied to other general cases like other bandit algorithms that implement Self-Normalized Bound for Vector-Valued Martingales from \citep{NIPS2011_e1d5be1c}. 
\begin{table*}[!t] 
  \caption{Notation of parameters}
  \centering
  \begin{adjustbox}{width=\textwidth}
  \begin{tabular}{*{4}{c} }
    \toprule
    \textbf{Notation} & \textbf{Definition} & \textbf{Dimension} &\textbf{Description} \\
    \midrule
    $y_t$ & $[y_{t,1},\ldots ,y_{t,N}]^T$
    & $N \times 1$ & The health conditions of all units in cycle $t$ \\
    & &  & where $y_{t, j} = 0$ when unit $j$ is not monitored \\
    
    $\textbf{X}_t$& $[x_{t,1},\ldots ,x_{t,N}]$ & $P \times N$ & The feature vectors of all units in cycle $t$\\
    & &  & where $x_{t, j} = \overrightarrow{0}$ when unit $j$ is not monitored \\
    
    $C_t$ & $[c_{1t},\ldots ,c_{Nt}]$ & $K \times N$ & The membership vectors of all units estimated in cycle $t$\\
    
    $Q_t$& $[q_{1t}, \ldots, q_{Kt}]$& $P \times K$ & The canonical model parameters estimated in cycle $t$\\
   \hline
    Notations used in \textbf{Step-1} of Section \ref{sub3.4.1} \\
   \hline
    
    $C_t^{(d)}$ &$diag(C_t)$
    & $NK \times N$ & The transformed $C_t$ \\
    
    $X_t$ &
    $ \textbf{X}_t \otimes I_{K} $
    & $Kp \times NK$& The transformed $\textbf{X}_t$\\
    
    $q_t$ & $vec(Q_t) $ &$Kp \times 1$& The transformed $Q_t$\\
    \hline
    Notations used in \textbf{Step-2} of Section \ref{sub3.4.1} \\
    \hline
    
    $E$ &  $[E_{1}, \cdots, E_{N}]$
    & $N \times N$ & Laplacian matrix \\
    
    $F$ &$\eta_2I_{N} + \lambda E$
    & $N \times N$ & The sum of the identity matrix $I_{N}$ \\
    &&& and Laplacian matrix $E$ \\
    
    $F_\otimes$& $F \otimes I_K$ & $NK \times NK$ & The Kronecker product of $F$ and $I_K$ \\
    
    $\Tilde{X_t}$ & $diag(\textbf{X}_t)$
    & $Np \times N$ & The transformed $\textbf{X}_t$ \\

    $\Tilde{Q}_t$ &$diag(Q_t,\ldots, Q_t)  F_\otimes^{-1/2}$
    & $Np \times NK$ & The transformed $Q_t$ \\
    
    $\Tilde{c}_t  $& $ F_\otimes^{1/2} vec(C_t)$ 
    & $NK \times 1$ & The transformed $C_t$ \\
    \bottomrule
  \end{tabular}
  \end{adjustbox}
  \label{table:1}
\end{table*}
\begin{figure}[!t]
\centering
\includegraphics[width=0.8\columnwidth]{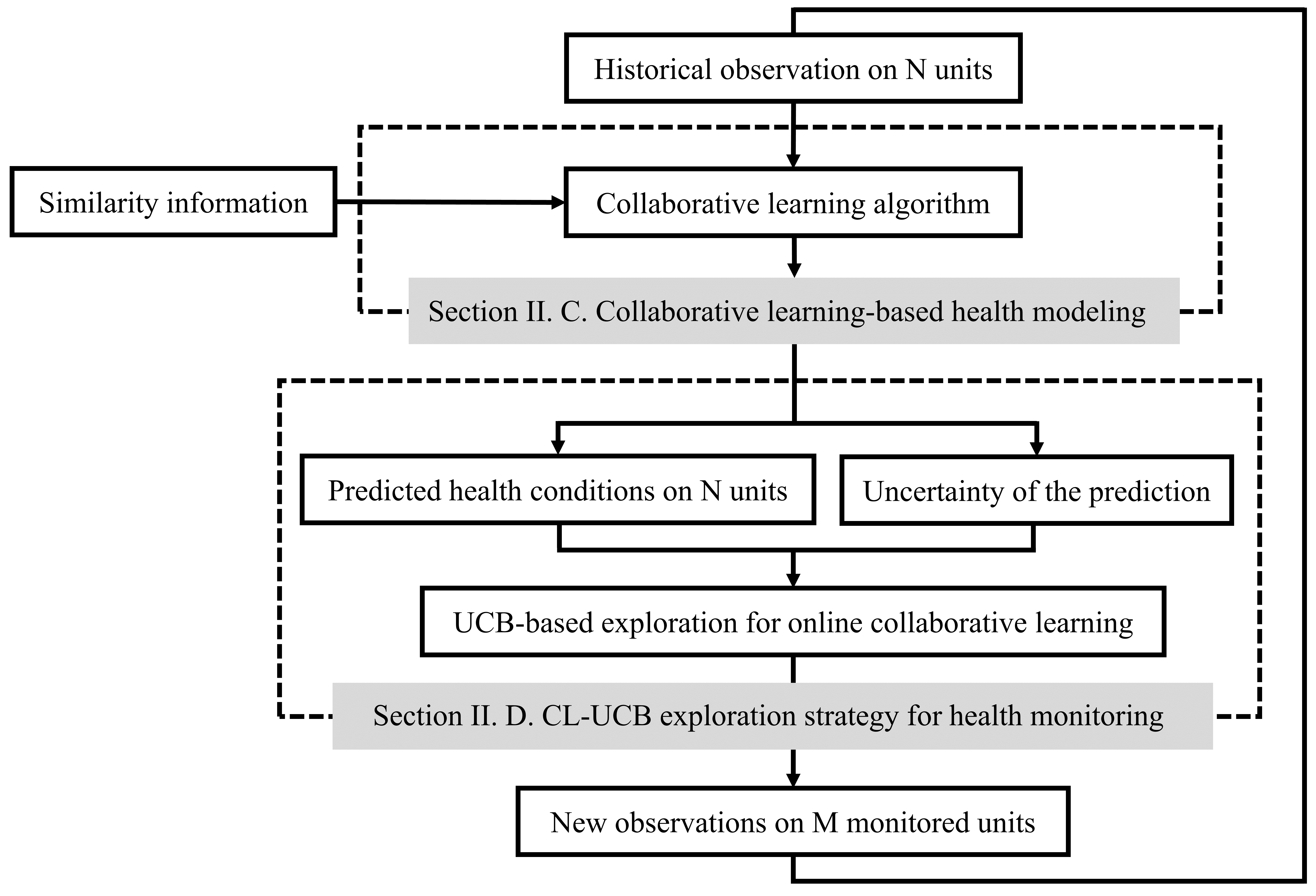}
\caption{The overall framework of the proposed Online Collaborative Learning (OCL) algorithm}
\label{fig:fig1}
\end{figure}
\section{Method}
All notations in this section are summarized in Table \ref{table:1}. 
\subsection{Problem Statement}
\label{sub3.1}
We investigate a population of $N$ units to be monitored over $T$ cycles. In each cycle, a feature vector is available for each unit to predict its health progression dynamics, denoted as $x_{t,i}$, $t\in\{1,\dots,T\}$, $i\in\{1,\dots,N\}$. If a unit is monitored in $t$th cycle, the unit's health condition is measured and denoted as $y_{t,i}$. The health monitoring system aims to select $M$ units for monitoring in each monitoring cycle so that units with most severe health conditions are maximally monitored. The monitoring decisions in cycle $t$ can be represented as a vector of binary variables $a_t$, with each element, $a_{t,i}$, indicates whether the $i$th unit is monitored $(a_{t,i} = 1)$ or not $(a_{t,i} = 0)$. The summation of binary variables in each cycle is constrained by the monitoring capacity, i.e. $\sum_{i = 1}^{N} a_{t,i} = M$. The reward of monitoring a unit with expected health condition $y_{t,i}$ is assumed to be known and characterized by a monotonic function $r(y_{t,i})$, which monotonically increase or decrease with respect the health outcome. The monotonic assumption ensures that monitoring a more severe unit is expected to lead to a higher reward than monitoring a healthy unit. Therefore, the reward of a monitoring strategy in cycle $t$ can be represented as:
\begin{equation} \label{eq:1}
    r_{a_t} = \sum_{i = 1}^{N}a_{t,i} r(y_{t,i})
\end{equation}

Clearly, with the knowledge of units' health conditions, the optimal monitoring strategy can be simplified as choosing the $M$ units with the highest monitoring rewards in each cycle. However, as the unit's health conditions and progression dynamics remain undisclosed, identifying the optimal monitoring strategy poses a challenge. To address this issue, we introduce an online collaborative learning approach. This method seamlessly integrates personalized health progression modeling, monitoring strategy design under limited resources, and online learning algorithms. In the following, we illustrate the model development under the scenario that reward of monitoring a unit is proportional to its health condition, i.e. $r(y_{t,i})=y_{t,i}$. However, the model is flexible to be extended to the other reward functions which are non-linear reward functions (e.g. logarithmic, probabilistic) and reward functions that depend on interactions between units. Also, the reward functions need to satisfy two general mild assumptions which are non-decreasing monotonicity and bounded smoothness \citep{chen2016combinatorial, qin2014contextual}. 

\subsection{Online Collaborative Learning (OCL) framework}
\label{sub3.2}
The proposed OCL method, as depicted in Figure \ref{fig:fig1}, encompasses two interconnected phases: health modeling and health monitoring. During the health modeling phase, historical data and prior information on units' similarities are utilized to estimate health progression of the units through a collaborative learning algorithm. In the health monitoring phase, an alternating least square algorithm is introduced to estimate the unknown parameters within the health modeling, which enables the prediction of their health conditions and the associated prediction uncertainties. Subsequently, an upper confidence bound (UCB)-based exploration strategy is designed to find the optimal monitoring strategy in the forthcoming cycle by integrating the predicted health conditions and their uncertainties. New observations collected on $M$ monitored units are used to update the model and monitoring strategy for the subsequent cycles.

\subsection{Collaborative learning-based health modeling}

\label{sub3.3}
To predict the health conditions of $N$ units in real time, their health progression dynamics need to be explicitly modeled. The health progression dynamic on each unit is represented by a distinct regression model that predicts the health condition in each cycle from the corresponding feature vector, i.e., $y_{t,i} = f_i(x_{t,i}, \beta_i) + \epsilon_i$. $f_i$ represents a parametric individualized regression model described by coefficients $\beta_i$ and $\epsilon_i$ is a random noise that follows the normal distribution. Parametric regression models have been popular tools in a wide range of system dynamics modeling problems, such as disease trajectory modeling and degradation modeling \citep{8169076}. It is also flexible to capture both linear and nonlinear dynamics in health progression by applying nonlinear transformations on the feature vectors \citep{filippi2010parametric}. The feature vectors can be defined using the basis functions of time or routinely observed information from units.

We leverage the idea of collaborative learning to capture the health progression dynamics for a population of dependent processes \citep{8169076}. The collaborative learning method approximates $N$ personalized health progression models by a small number of canonical models that represent the typical mechanisms in health progression and span the parametric space of personalized health progression models. Each unit's health progression model is represented by the ensemble of canonical models with a personalized membership vector. Let  $g^{1}(x),\dots,g^{K}(x)$ represent $K$ canonical models in the population, the health progression dynamic for unit $i$ can be estimated by the weighted combination of the predictions of from $K$ canonical models, i.e. $f_i (x) = \sum_{k} c_{ik} g^{k}(x)$; $c_{ik}$ is the element in unit $i$’s membership vector $c_i$. For simplicity, here, we focus on a linear regression case, where each canonical model can be described by a linear regression model with parameters $q_k$ i.e., $g^k (x) = x^T q_k$. With the linear regression assumption, the health progression dynamic for unit $i$ can be approximated as $f_i(x)=\sum_{k} c_{ik}x^T q_k$. The collaborative learning method under linear assumption can be regarded as a matrix factorization problem which decomposes the coefficients of $N$ linear regression models to two low-rank matrices, denoted as $\beta_i=Qc_i$. $Q = [q_1,\ldots, q_K] \in \mathbb{R}^{p \times K}$ represents the parameters in canonical models. In this paper, we assume the canonical models and membership vectors are unconstrained and let data freely determine the inherent group structure. It is notable that the collaborative learning method is also flexible to capture the nonlinear dynamics in health progression by applying nonlinear transformations on the feature vectors, such as the Generalized Linear Model-UCB (GLM-UCB) algorithm developed in \citep{filippi2010parametric}.

The collaborative learning method can further exploit the similarity information between units to refine model learning by assuming that units sharing similarities tend to follow similar health progression dynamics. It treats the similarities between units as prior knowledge, which can be sourced from domain experts or units' profiles on risk factors. This similarity information is then employed to regularize the learning of personalized membership vectors. Given the similarity information between units and 
the dependency between units is taken into account, a graph laplacian matrix $E \in \mathbb{R}^{N \times N}$ is used to represent the similarity information between $N$ units \citep{DBLP:journals/corr/Cesa-BianchiGZ13}.
The collaborative learning method is formulated as a nonconvex optimization problem in the following, which simultaneously minimizes the training error on historical data and the variations between similar units: 
\begin{equation} \label{eq:2}
    \min_{C, Q} \sum_{t} \sum_{i} \left\| x_{t, i}Qc_i - y_{t, i}\right\|_2^{2} +\eta_1\left\|Q\right\|_F^{2}
    +\eta_2\left\|C\right\|_F^{2}
    + \lambda Tr(C^TEC)
\end{equation} 
where $C = [c_1,\ldots, c_N]$. L2 regularizations are used on $Q$ and $C$ to control the scale of unknown parameters and enforce unique solution with the tuning parameters $\eta_1$ and $\eta_2$. $\lambda$ is the tuning parameter to control the effect of the similarity information on the parameter estimation.

\subsection{CL-UCB exploration strategy for health monitoring}
\label{sub3.4}
As outlined in Section \ref{sub3.1}, our main problem is to 
estimate the health progression dynamics for $N$ units to predict their health conditions in real-time, such that $M$ out of $N$ units with the most severe health conditions are prioritized to be monitored.
This section provides the details of how to integrate the collaborative learning approach described in Section \ref{sub3.2} to the health monitoring phase. The key idea in health monitoring phase is to select $M$ units that best balance the objectives of monitoring units with severe predicted health conditions (exploitation) and monitoring the units with high uncertainty in health predictions (exploration). Therefore, it is critical to obtain the uncertainty in health predictions. To achieve this goal, the collaborative learning model is first transformed and efficiently solved by an alternating least square algorithm in Section \ref{sub3.4.1}, which provides closed form solutions for each step of parameter updates. Then the confidence bounds of parameters estimated from the alternating least square algorithm are derived theoretically in Section \ref{sub3.4.2}. Finally, a novel collaborative learning-based UCB (CL-UCB) exploration strategy is proposed to online update the collaborative learning model and inform monitoring strategy design based on the units' predicted health conditions and their confidence bounds.

\subsubsection{Alternating Least Square Algorithm to Estimate Collaborative Regression Models.}
\label{sub3.4.1}
To estimate the parameters in collaborative learning, an alternating least square algorithm is proposed that minimizes the objective function in Equation \ref{eq:2}. The proposed alternating least square algorithm alternatively updates the model parameters using two main steps. It estimates the canonical models by fixing the membership vectors in \textbf{Step-1} and then updates the membership vectors with fixed canonical models in \textbf{Step-2}.

\textbf{Step-1: When the membership vectors are fixed at $C_t$, the estimation of canonical models can be obtained by solving a least-square estimation problem.} 
In cycle $t$, the health conditions of all units can be represented as $y_t = [y_{t,1},\ldots ,y_{t,N}]^T \in \mathbb{R}^{N \times 1}$, and the corresponding feature matrix $\textbf{X}_t = [x_{t,1},\ldots ,x_{t,N}] \in \mathbb{R}^{p \times N}$ where $x_{t,j} = \overrightarrow{0}$ and $y_{t,j} = 0$ for units that are not monitored. 
We transform the feature matrix and membership vectors as $X_t = \textbf{X}_t \otimes I_{K} \in \mathbb{R}^{Kp \times NK}$ and $C_{t}^{(d)} = diag(C_{t}) \in \mathbb{R}^{NK \times N}$, where $diag(C_t)$ is the operation that returns the diagonal matrix whose diagonal blocks equal to the fixed membership vectors in cycle $t$ and non-diagonal blocks are set to zero,
By representing the parameters in canonical models as a $Kp \times 1$ vector, $q = vec(Q) = [q_1^T,\cdots,q_K^T]^T$ where $vec(\cdot)$ is the vectorize operation, the problem of estimating the canonical models under fixed membership vectors can be reformulated as a least-square estimation problem as follows: 
\begin{equation} \label{eq:3}
     \min_{q} \sum_{t^{\prime}=1}^{t} \left\| C_{t}^{(d)T}X_{t^{\prime}}^Tq - y_{t^{\prime}} \right\|_2^{2} + \eta_1 \Vert q\Vert_2^2
\end{equation}
The closed-form estimation of $q$ with respect to Equation \ref{eq:3} can be achieved by $\hat{q_t} = A_t^{-1} b_t$, where
\begin{align}
    A_t ={}& \sum_{t^{\prime} = 1}^{t} X_{t^{\prime}}C_{t}^{(d)}C_{t}^{(d)T}X_{t^{\prime}}^T  + \eta_1 I_{Kp} \label{eq:4}\\
    b_t ={}& \sum_{t^{\prime} = 1}^{t} X_{t^{\prime}}C_{t}^{(d)}y_{t^{\prime}} \label{eq:5}
\end{align}



\textbf{Step-2: When the canonical models are fixed at $Q_t$, the estimation of membership vectors can be formulated as another least square estimation problem.} 
Similar to the approach in \citep{DBLP:journals/corr/Cesa-BianchiGZ13}, we use the matrix $F = I_{N} + \lambda E$ to represent the regularization, where $I_{N}$ is the $N \times N$ identity matrix and $\lambda$ is the tuning parameter to control the effect of the regularization term on the parameter estimation. $F$ is transformed into $F_\otimes \in \mathbb{R}^{NK \times NK}$ where $F_\otimes= F \otimes I_K$. The feature matrix and canonical model are rewritten as $\Tilde{X_{t}} = diag(\textbf{X}_t) \in \mathbb{R}^{Np \times N}$ and $\Tilde{Q}_{t} =  diag(Q_t,\ldots, Q_t) F_\otimes^{-1/2} \in \mathbb{R}^{Np \times NK}$ respectively.
Denoting the parameters of membership vectors as $\Tilde{c}  = F_\otimes^{1/2} vec(C) \in \mathbb{R}^{NK \times 1}$, the regularized objective function can be efficiently represented as Equation \ref{eq:6}, which can be easily solved with closed-form solution as follows.
\begin{equation} \label{eq:6}
    \min_{\Tilde{c}} \sum_{t^{\prime}=1}^{t} \left\| \Tilde{X_{t^{\prime}}}^T \Tilde{Q}_{t}\Tilde{c} - y_{t^{\prime}} \right\|_2^{2} + \eta_2\Vert \Tilde{c} \Vert_2^2
\end{equation}
The closed-form estimation of $\Tilde{c}$ with respect to Equation \ref{eq:6} can be achieved by $\hat{\Tilde{c_t}} = D_t^{-1} d_t$ in which,
\begin{align}
    D_t ={}& \sum_{t^{\prime}=1}^{t}  \Tilde{Q}_{t}^T\Tilde{X_{t^{\prime}}}\Tilde{X_{t^{\prime}}^T}\Tilde{Q}_{t} + \eta_2I_{NK} \label{eq:7}\\
    d_t ={}& \sum_{t^{\prime}=1}^{t} \Tilde{Q}_{t}^T\Tilde{X_{t^{\prime}}} y_{t^{\prime}} \label{eq:8}
\end{align}


\begin{algorithm}[t]
\caption{Alternating Least Square Algorithm for Solving Collaborative Learning}\label{alg:alg1}
\begin{algorithmic}[1]
\STATE \textbf{Input:} $\{X_{t, i}, \Tilde{x}_{t, i}, y_{t, i}\} 
  , \forall i \in \Tilde{S}_t$ that contains $M$ arms with the highest $\hat{u}_{t, i}$ 
\STATE Initialize $A_{t+1} = A_t, b_{t+1} = b_t, D_{t+1} = D_t, d_{t+1} = d_t, \hat{q}_{t+1} = \hat{q}_{t}, \hat{\Tilde{c}}_{t+1} = \hat{\Tilde{c}}_{t}$
\WHILE{not converge}
\STATE Transform $\hat{q}_{t+1}$ to $\hat{\Tilde{Q}}_{t+1}$
\FOR{$i \in \Tilde{S}_t$}
    \STATE $A_{t+1} \gets A_{t+1} + $ 
    $ X_{t, i} \hat{\Tilde{c}}_{t+1} \hat{\Tilde{c}}_{t+1}^T  X_{t, i}^T$ 
    \STATE $b_{t+1} \gets b_{t+1} +  X_{t, i} \hat{\Tilde{c}}_{t+1}y_{t, i}$
    \STATE $D_{t+1} \gets D_{t+1} + \hat{\Tilde{Q}}_{t+1}^T\Tilde{x}_{t, i}\Tilde{x}^T_{t, i}\hat{\Tilde{Q}}_{t+1}$
    \STATE $d_{t+1} \gets d_{t+1} + \hat{\Tilde{Q}}_{t+1}^T\Tilde{x}_{t, i}
    y_{t, i}$
\ENDFOR
    \STATE $\hat{q}_{t+1} \gets A_{t+1}^{-1}b_{t+1}$
    \STATE $\hat{\Tilde{c}}_{t+1} \gets D_{t+1}^{-1}d_{t+1}$
\ENDWHILE
\STATE \textbf{Output:} ${\hat{q}_{t+1}, \hat{\Tilde{c}}_{t+1}}$
\end{algorithmic}
\label{alg1}
\end{algorithm}
The procedure of the proposed alternating least square algorithm for estimating the collaborative learning model is summarized in Line 2 - Line 12 of \textbf{Algorithm \ref{alg:alg1}}. 

\subsubsection{CL-UCB Exploration Strategy for Online Health Modeling and Monitoring.}
\label{sub3.4.2}
The canonical models and membership vectors estimated from the alternating least square algorithm provide the predicted health condition for all $N$ units in the next cycle $t+1$ by $\hat{y}_{t+1} = \hat{\Tilde{c}}^{T}_{t+1}X_{t+1}^{T}\hat{q}_{t+1}$. It represents the model’s
current best knowledge about disease progression and therefore serves for exploitation purposes. However, only selecting units that have the highest predicted health conditions is not sufficient in designing an optimal monitoring strategy since the uncertainty in prediction and these units cannot guarantee to be optimally selections for the entire monitoring period. Thus, it is critical to design the exploration part in the monitoring strategy which allows uncertain units to be explored to gain more information and more confidence in selection. To balance the trade-off between exploitation which is monitoring high-risk processes based on existing knowledge and exploration which is gaining the knowledge of monitoring uncertain processes, the upper confidence bound (UCB) principle is commonly used to estimate the confidence of units' predicted health conditions and inform the exploration of most uncertain units \citep{auer2002using}. 
\begin{algorithm}[t]
\caption{Collaborative Learning UCB (CL-UCB)}\label{alg:alg2}
\begin{algorithmic}[1]
\STATE \textbf{Input:} $\eta_1, \eta_2, \lambda, \alpha^q, \alpha^{\Tilde{c}}, M$
\STATE Initialize $A_0 \gets \eta_1 I_{Kp}$, $b_0 \gets 0$, $D_0 \gets \eta_2I_{NK}$, $d_0 \gets 0$
\FOR{$t \gets 1,\ldots,T$}
    \STATE Observe features of all units $i \in [N]: x_{t, i} \in \mathbb{R}^{p \times 1}$ and transform to a sparse matrix $X_{t, i} \in \mathbb{R}^{Kp \times NK}$ and a sparse vector $\Tilde{x}_{t, i} \in \mathbb{R}^{Np \times 1}$, and $\hat{q}_{t}$ to $\hat{\Tilde{Q}}_t$
    \FOR{all $i \in [N]$}
        \STATE $\hat{u}_{t, i} \gets \hat{\Tilde{c}}_{t}^T X_{t, i}^T \hat{q}_{t} + \alpha^{\Tilde{c}} \sqrt{\Tilde{x}_{t, i}^T\hat{\Tilde{Q}}_{t}D_{t}^{-1} \hat{\Tilde{Q}}_{t}^T\Tilde{x}_{t, i}}+ \alpha^q \sqrt{\hat{\Tilde{c}}_{t}^T X_{t, i}^T A_{t}^{-1} X_{t, i}\hat{\Tilde{c}}_{t}}$
    \ENDFOR
    \STATE Use Algorithm \ref{alg:alg1} to solve the minimization problem in Equation \ref{eq:2}.
\ENDFOR
\end{algorithmic}
\label{alg2}
\end{algorithm}
We propose a new algorithm called Collaborative Learning-UCB (CL-UCB) to optimally allocate the monitoring resources that balances the trade-off between exploitation and exploration in online modeling and monitoring. It leverages the upper confidence bound (UCB) principle to measure the uncertainty of estimated parameters, i.e., the upper confidence bounds of $\Vert\hat{q_t} - q^*\Vert_{A_t}$ and $\Vert\hat{\Tilde{c_t}} - \Tilde{c}^*\Vert_{D_t}$ where $q^*$ and $\Tilde{c}^*$ are the ground-truth parameters. Then, it integrates the predicted rewards with their uncertainty to balance the exploitation-exploration tradeoff. The overview of CL-UCB is illustrated in \textbf{Algorithm \ref{alg:alg2}}. Based on the closed-form solution in the alternating least square algorithm, the confidence bound of $\hat{q_t}$ and $\hat{\Tilde{c_t}}$ can be obtained by the \textbf{Lemma \ref{lemma1}} below.
\begin{lemma}
\label{lemma1}
When the Hessian matrix of the objective function in Equations \ref{eq:3} and \ref{eq:6} are positive definite at the optimizer $q^*$ and $\Tilde{c^*}$ with proper initialization, for any $\epsilon_1 \geq 0$, $\epsilon_2 \geq 0$, $\Vert X_t\Vert_2 \leq S$, $\Vert q_t\Vert_2 \leq L$, $\Vert C_t\Vert_2 \leq P$, and for any $\delta \geq 0$, with probability at least 1 - $\delta$, the estimation errors of the canonical models and the membership vectors obtained from \textbf{Algorithm \ref{alg:alg1}} are upper bounded by:
\begin{equation}
    \Vert\hat{q_t} - q^*\Vert_{A_t} \leq{} \sqrt{Kp\ln(\frac{\eta_1 Kp + tS^2P^2}{\eta_1 Kp\delta})}
    + \frac{2SPL}{\sqrt{\eta_1}}\frac{(v_1+\epsilon_1)(1-(v_1+\epsilon_1)^t)}{1-(v_1+\epsilon_1)} + \sqrt{\eta_1}L \label{eq:9}
\end{equation}
\begin{align}
    \Vert\hat{\Tilde{c_t}} - \Tilde{c}^*\Vert_{D_t} &\leq{} \sqrt{NK\ln(\frac{\eta_2NK + S^2L^2 \sum_{t^{\prime} = 1}^{t} \sum_{j = 1}^{NK} (f^{-1}_\otimes)_{j,j}}{\eta_2 NK\delta})} + \frac{2SPL}{\sqrt{\eta_2}}\frac{(v_2+\epsilon_2)(1-(v_2+\epsilon_2)^t)}{1-(v_2+\epsilon_2)} \notag\\
    &+ \eta_2 L(c_1,\ldots,c_n) \label{eq:10}
\end{align}
in which $0 < v_1 < 1$, $0 < v_2 < 1$ 
\end{lemma}
The detailed proof of \textbf{Lemma \ref{lemma1}} is provided in the Appendix. From \textbf{Lemma \ref{lemma1}}, denote $\alpha^q$ and $\alpha^{\Tilde{c}}$ as the upper bounds of $\Vert\hat{q_t} - q^*\Vert_{A_t}$ and $\Vert\hat{\Tilde{c_t}} - \Tilde{c}^*\Vert_{D_t}$ respectively, we formulate a \textbf{Proposition \ref{proposition1}} which defines the upper confidence bound of the predicted health condition and serves as the uncertainty of the predicted health condition.
\begin{proposition}
\label{proposition1}
 With the probability at least $1-\delta$, 
\begin{equation}
    \lvert y_{t,i}^* - \hat{y}_{t, i} \rvert \leq{}  \alpha^{\Tilde{c}} \sqrt{\Tilde{x}_{t, i}^T\hat{\Tilde{Q}}_{t}D_{t}^{-1} \hat{\Tilde{Q}}_{t}^T\Tilde{x}_{t, i}} + \alpha^q \sqrt{\hat{\Tilde{c}}_{t}^T X_{t, i}^T A_{t}^{-1} X_{t, i}\hat{\Tilde{c}}_{t}} \label{eq:11}
\end{equation} 
\end{proposition}
The detailed proof of Equation \ref{eq:11} is shown in the Appendix. However, selecting units only with the highest uncertainty (i.e., exploration) from Equation \ref{eq:11} is not optimal since it does not consider maximizing their predicted health condition (i.e., exploitation). To balance the exploitation-exploration tradeoff, a unit's CL-UCB score is proposed for monitoring strategy design, which combines its real-time predicted health condition and the associated prediction uncertainty. Denote $\alpha^q$ and $\alpha^{\Tilde{c}}$ as the upper bounds of $\Vert\hat{q_t} - q^*\Vert_{A_t}$ and $\Vert\hat{\Tilde{c_t}} - \Tilde{c}^*\Vert_{D_t}$ obtained from \textbf{Lemma \ref{lemma1}} respectively, the CL-UCB score of unit $i$ is defined as follow:
\begin{equation}
   \hat{u}_{t, i} = \hat{\Tilde{c}}_{t}^T X_{t, i}^T \hat{q}_{t} + \alpha^{\Tilde{c}} \sqrt{\Tilde{x}_{t, i}^T\hat{\Tilde{Q}}_{t}D_{t}^{-1} \hat{\Tilde{Q}}_{t}^T\Tilde{x}_{t, i}} + \alpha^q \sqrt{\hat{\Tilde{c}}_{t}^T X_{t, i}^T A_{t}^{-1} X_{t, i}\hat{\Tilde{c}}_{t}} \label{eq:12}
\end{equation} 
Equation \ref{eq:12} illustrates the CL-UCB score $\hat{u}_{t, i}$ which consists of two parts.The first term represents the predicted health condition of unit $i$. The second and third terms represent the prediction uncertainty due to the canonical models and membership vector estimations respectively. To allocate the resources to both high-risk and uncertain processes, we select $M$ units that have the highest CL-UCB score to be monitored in the next cycle. The procedure of CL-UCB is summarized in \textbf{Algorithm \ref{alg:alg2}}.

\subsection{Regret Analysis}
\label{sub3.5}
In this section, we provide a theoretical analysis of the cumulative regret of the Collaborative Learning UCB (CL-UCB) algorithm. Based on \textbf{Lemma \ref{lemma1}}, the alternating least square-based parameter estimation satisfies the two inequalities shown in Equation \ref{eq:9} and \ref{eq:10} and it contributes to the final regret of the algorithm. Denote the optimal monitoring strategy as $a_t^*$, the corresponding optimal reward can be represented as $r_{a_t^*} = \sum_{i} a_t^*y_{t,i}$.The difference between the proposed algorithm’s total reward and the total reward of the optimal strategy is called cumulative regret, which is defined as:
\begin{equation} \label{eq:13}
    R(T) = \sum_{t = 1}^{T}R_t = \sum_{t = 1}^{T}(\sum_{i} a_t^*y_{t,i} - \sum_{i} a_ty_{t,i})
\end{equation}
\textbf{Theorem \ref{theorem1}} provides the upper regret bound of the CL-UCB algorithm, which analyzes the quality of the proposed monitoring strategy theoretically.

\begin{theorem}
\label{theorem1}
Based on a proper initialization of the alternating least square algorithm, with the probability at least 1 - $\delta$, the cumulative regret of the CL-UCB algorithm is upper bounded by: 
\begin{align*}
    R(T) &\leq 2\alpha^{\Tilde{c}}\sqrt{2TNK\ln{(1 + \frac{S^2L^2\sum_{t^{\prime} = 1}^{t} \sum_{j = 1}^{NK} (f^{-1}_\otimes)_{j,j}}{\eta_2NK})}} + 2\alpha^q\sqrt{2TKp\ln{(1 + \frac{TS^2P^2}{\eta_1 Kp})}}  \notag\\ 
    &+ \frac{2mv^2(1-v^{2T})}{1-v^2} 
\end{align*}
\end{theorem}
According to \textbf{Theorem \ref{theorem1}}, the term $\sum_{t^{\prime} = 1}^{T} \sum_{j = 1}^{NK} (f^{-1}\otimes)_{j,j}$ plays a significant role in the upper regret bound of CL-UCB, highlighting the impact of the similarity information. First, if $F$ is an identity matrix, meaning that the proposed method solely considers the latent group structure without similarity information, then $\sum_{t^{\prime} = 1}^{t} \sum_{j = 1}^{NK} (f^{-1}_\otimes)_{j,j} = NKT$and leads to the upper regret bound of $O(NK\sqrt{T}\ln{T} + (1-c^T_{\Tilde{c}})\sqrt{T\ln{T}} + Kp\sqrt{T}\ln{\frac{T}{Kp}} + (1-c^T_{q})\sqrt{T\ln{\frac{T}{Kp}}})$ where $c_{\Tilde{c}}$ and $c_q$ are the constants between 0 and 1. If $K \ll p$, capturing the low-rank structure in health progression dynamics using canonical models and membership vectors can improve the upper regret bound compared to modeling the units independently in LinUCB which has the regret bound of $O(Np\sqrt{T}\ln{T})$. Second, when the similarity matrix is fully connected, i.e., $(f_\otimes)_{j,j} = N$, then $\sum_{t^{\prime} = 1}^{t} \sum_{j = 1}^{NK} (f^{-1}_\otimes)_{j,j} = \frac{2NKT}{N+1}$. This leads to the upper regret bound of $O(NK\sqrt{T}\ln{\frac{T}{N}} + (1-c^T_{\Tilde{c}})\sqrt{T\ln{\frac{T}{N}}} + Kp\sqrt{T}\ln{\frac{T}{Kp}} + (1-c^T_{q})\sqrt{T\ln{\frac{T}{Kp}}})$ where $c_{\Tilde{c}}$ and $c_q$ are the constant between 0 and 1. Compared to LinUCB, the proposed method which allows all units to have homogeneous influence to each other exhibits a lower regret bound by $\frac{1}{N}$. Also, compared to the algorithm only using similarity information, i.e. GOB.Lin algorithm \citep{DBLP:journals/corr/Cesa-BianchiGZ13} which has the regret bound of $O(Np\sqrt{T}\ln{\frac{T}{N}})$, it indicates that by utilizing the low-rank structure, the proposed CLUCB achieves a regret reduction since $K \ll p$.

\subsection{Empirical Issues of Implementing the Algorithm}
There are some empirical issues with implementing the algorithm. First, to estimate the unknown parameters in each cycle, in theory, we need to update the parameters for a large number of iterations in \textbf{Algorithm \ref{alg:alg1}} to guarantee convergence. However, the computational cost of \textbf{Algorithm \ref{alg:alg1}} can be high in practice. To save the computational costs, we update the parameters with a smaller number of iterations in each cycle. Second, there are several tuning parameters, such as $L2$ regularization parameters $\eta_1, \eta_2,$ similarity regularization parameter $\lambda$, and exploration parameters $\alpha^q, \alpha^{\Tilde{c}}$, that affect the effectiveness of the proposed method. In our experiment, to save computation costs and focus more on the effect of exploration parameters in different algorithms, we fix all $L2$ regularization parameters $\eta_1 = \eta_2 = 0.3$, $\lambda = 0.01$ so that the unit exploration is not too biased by provided similarity information, i.e. graph laplacian matrix $E$, and empirically tune $\alpha^q$ and $ \alpha^{\Tilde{c}}$ to obtain the optimal performance of the proposed method. The selections of tuning parameters are based on the noisiness of the data and the complexity of the latent structure. If the data is noisy and the number of latent structure ($K$) is large, the tuning parameter for latent structure $\alpha^q$ prefers high values since the bandit algorithm requires more exploration to gain information about units. However, trade-off associated with increased exploration is higher cumulative regret since the model needs to spend more cycles to learn more complicated latent structure. For the estimation of membership vectors, the similarity regularization parameter $\lambda$ prefers high values to encourage encourage membership vectors to be more similar if the similarity between two units is larger. 
\begin{figure}[!t]
\centering
\includegraphics[width=\columnwidth]{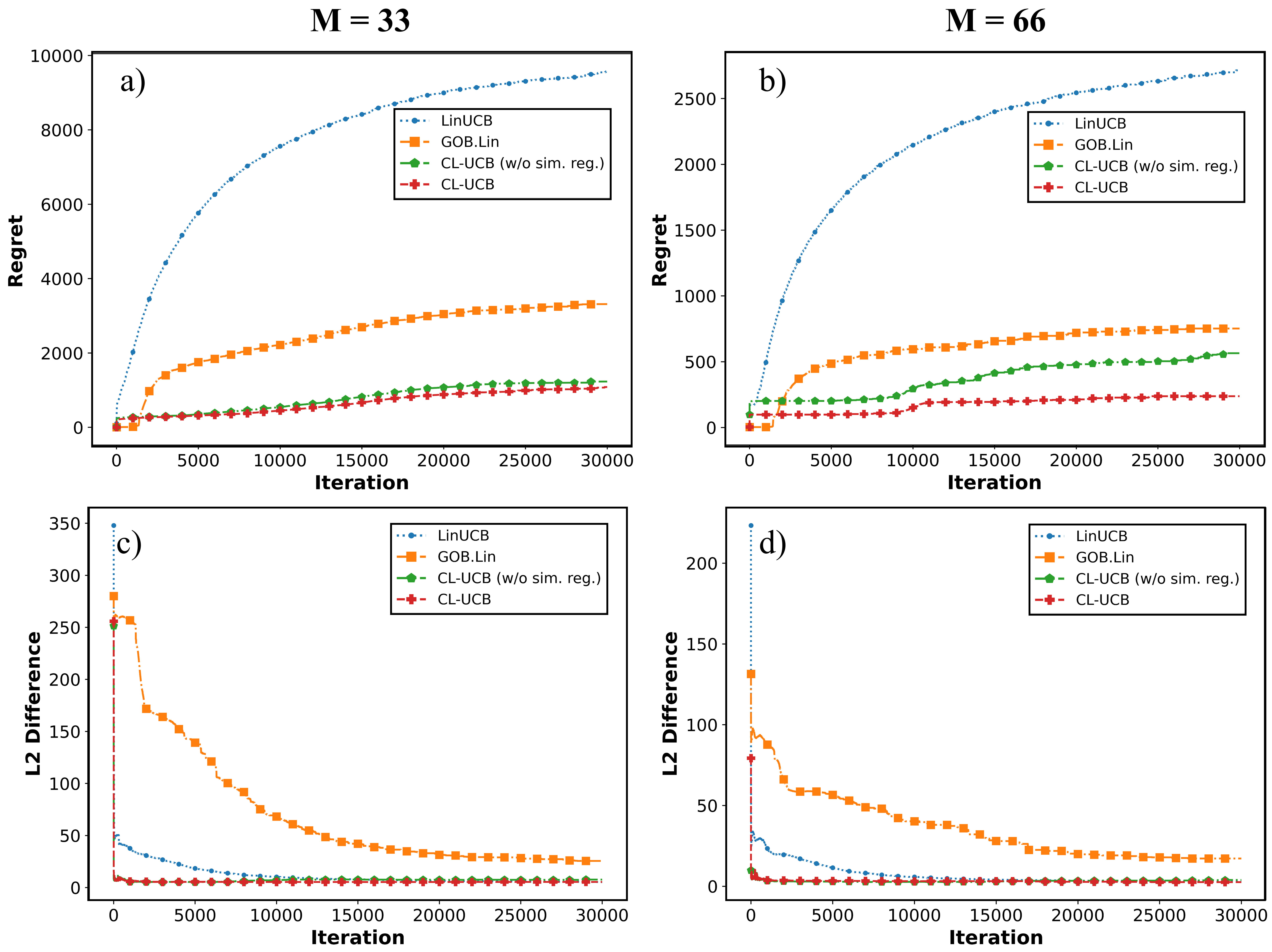} 
\caption{a)-b) The convergence of cumulative regret, c)-d) The accuracy of bandit parameter estimation}
\label{fig:fig2}
\end{figure}

\section{Experiments}
The proposed algorithm is tested in the context of cognitive degradation monitoring of Alzheimer's Disease (AD) in older populations. To evaluate both monitoring effectiveness and the accuracy of modeling learning, a simulation model of cognitive degradation monitoring is first built and simulation study is conducted. Then the algorithm is tested on a real-world population. A cognitive measure, Mini-Mental State Examination (MMSE) \citep{arevalo2015mini}, is used as the outcome of cognitive monitoring in all experiments. To model the progression of MMSE, we consider a common type of degradation model that uses the time or its polynomial basis functions as predictors to predict the cognitive status \citep{10.1093/cercor/bhh003, bartzokis2004heterogeneous}. The cognitive degradation process of the unit $i$ is formulated by the following \citep{biesanz2004role}.
\begin{equation} \label{eq:sim}
y_{it} = \beta_{i0} + \beta_{i1}t + \beta_{i2}t^2+\epsilon_{it}
\end{equation}
where $y_{it}$ is the MMSE measurement of the unit $i$ at time point $t$, $\beta_{i0}$ represents the unit's initial cognitive status, $\beta_{i1}$ and $\beta_{i2}$  indicate the velocity and acceleration of the unit's cognitive degradation respectively and $\epsilon_{it}$ is the Gaussian noise. To capture the typical cognitive degradation mechanisms in the population and dependencies among units, the proposed method is applied to estimate the coefficients in cognitive degradation model through sequentially observed data and inform the monitoring resources allocation. 
Two monitoring scenarios are considered to represent different levels of resource constraints in real world: the high-constraint scenario which only monitors patients with severe cognitive conditions ($M=33\%$ of the total population) and the relaxed-constraint scenario which monitors patients with both severe and mild cognitive impairments ($M=66\%$ of the total population).  All experiments continuously monitor the populations over 30000 cycles ($T = 30000$).

The proposed algorithm (CL-UCB), the proposed algorithm without similarity regularization (CL-UCB w/o similarity), and two benchmark MAB algorithms are compared. The benchmark MAB algorithms include LinUCB \citep{li2010contextual}, which does not account for the dependency between processes and GOB.Lin \citep{DBLP:journals/corr/Cesa-BianchiGZ13}, which solely considers the similarities between processes. LinUCB estimates the coefficients in each polynomial regression model independently via L2-regularized least square estimation. GOB.Lin leverages the similarities between units in the parameter estimation through the graph Laplacian regularization. However, neither algorithm explicitly exploits the latent structure present in health progression. The cumulative regret defined in Eq.\ref{eq:13} is used to compare the performance of different monitoring algorithms in both simulation and real-world studies. In the simulation study, the L2-norm difference between estimated parameters and true parameters is also used to evaluate the model learning accuracy.
\subsection{Simulation Studies}
In the simulation studies, we simulate the cognitive measures of $100$ units ($N=100$) over time from the degradation model presented in Equation \ref{eq:sim}. To incorporate the latent group structure in cognitive degradation, the coefficients in cognitive degradation models are constructed from a set of latent groups and individualized membership vectors, denoted as $\beta_i = Qc_i$.
The parameters in latent groups ($Q$) are randomly generated to represent different types of cognitive degradation patterns. Three types of cognitive decline are used in this study ($K=3$), which represent the normal aging cognitive decline, mild cognitive impairment and patients with AD \citep{MUELLER2005869}. To simulate the health progression dynamics of each unit from these latent groups, the membership vector of each unit ($c_i$) need to be simulated. Each membership vector has three elements that represent the difference among units. As the membership vectors should be able to cluster the units to three latent groups, they are randomly generated from a mixture Gaussian distributions with three zero-mean components and their respective prior probabilities are estimated from the real data.:
\[ F_1(c) \sim N(0, \begin{bmatrix} \sigma^2 & 0 & 0\\ 0 & 1 & 0\\ 0 & 0 & 1 \end{bmatrix}),F_2(c) \sim N(0, \begin{bmatrix} 1 & 0 & 0\\ 0 & \sigma^2 & 0\\ 0 & 0 & 1 \end{bmatrix}),F_3(c) \sim N(0, \begin{bmatrix} 1 & 0 & 0\\ 0 & 1 & 0\\ 0 & 0 & \sigma^2 \end{bmatrix})\]
The covariance matrix in each component is a diagonal matrix with $k$th diagonal element equals to $\sigma^2$ and the other elements equal to 1. It controls the significance of low-rank canonical structure. A larger value of $\sigma^2$ indicates a more distinct difference between canonical models. In this experiment, we set $\sigma^2$ at 100 to simulate three distinct groups. The random noises are generated from a standard normal distribution. The similarities between patients are estimated by calculating the cosine similarity of their membership vectors, i.e. $w_{ij} = c_{i}^Tc_{j}$.  All hyperparameters are carefully tuned based on the cumulative regret. To find the initial value of $c_i$, we use a k-means clustering algorithm to learn the centroid vectors from the similarity matrix $W$ and assign the initial value to $c_i$ based on the cluster it belongs to. Then, we apply the initial value of $c_i$ in the updating part from Algorithm \ref{alg:alg1} to initialize $Q$.
\begin{figure}[!t]
\centering
\includegraphics[width=0.7\columnwidth]{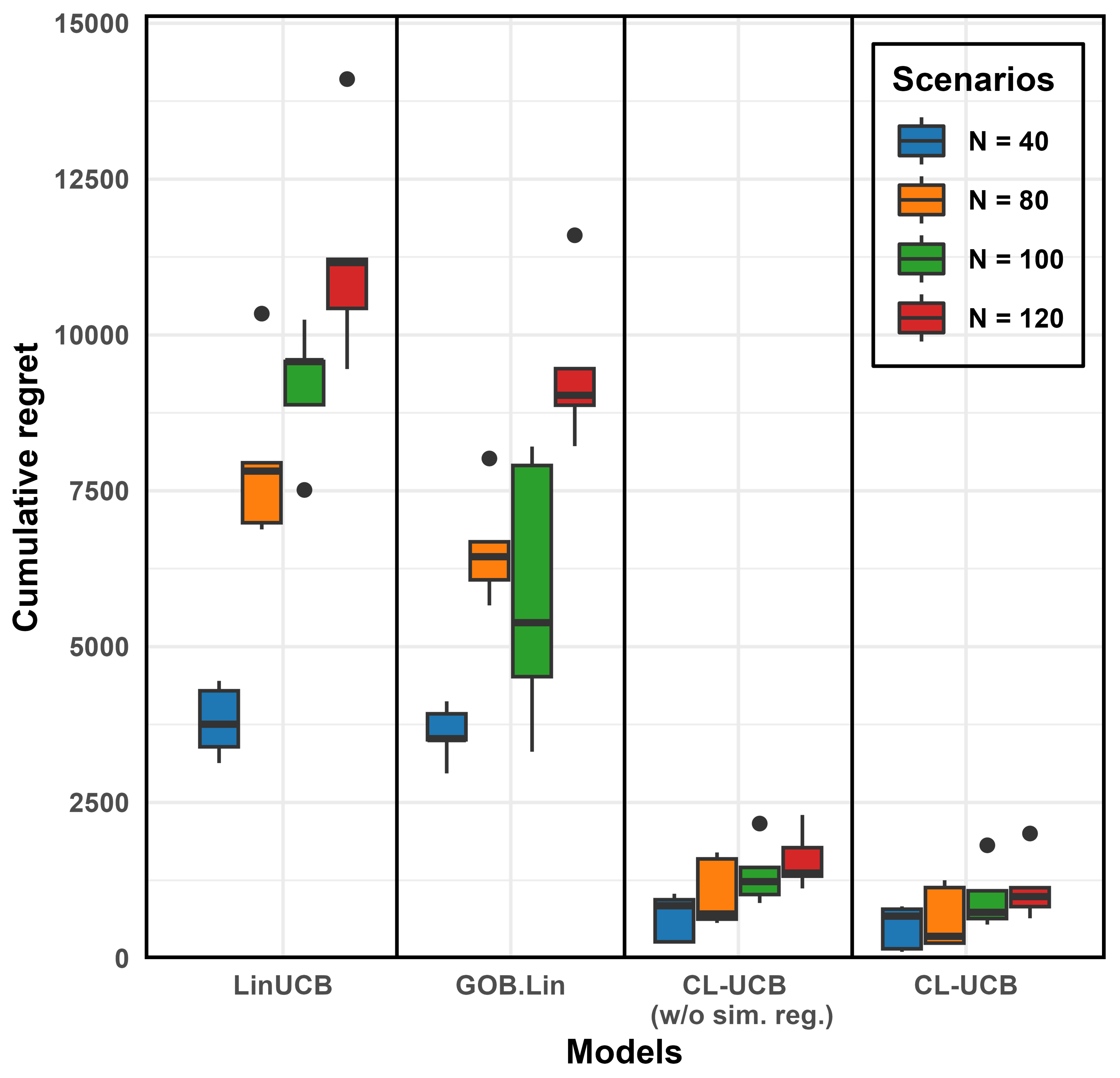} 
\caption{Box plot of cumulative regret with different bandit size $N$ }
\label{fig:box1}
\end{figure}
\begin{figure}[!t]
\centering
\includegraphics[width=0.7\columnwidth]{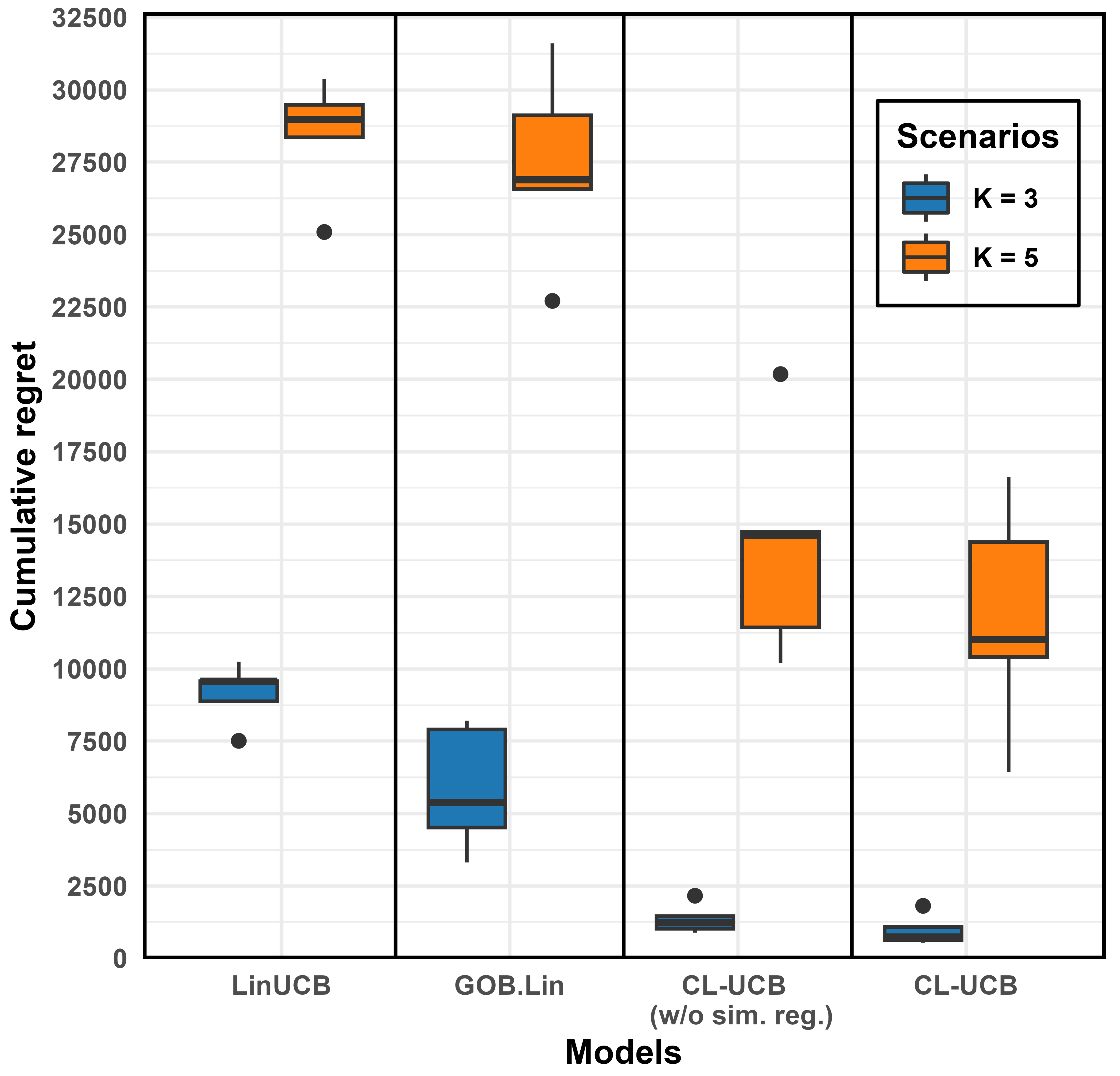} 
\caption{Box plot of cumulative regret with different number of latent structure $K$}
\label{fig:box2}
\end{figure}

In this study, there are two scenarios to evaluate the performance of all monitoring algorithms: the high-constraint scenario and the relaxed-constraint scenario. The high-constraint scenario represents the case where we monitor the entire group of patients with severe conditions while the relaxed-constraint scenario represents the case where we monitor the entire group of patients with severe and mild conditions. The cumulative regrets of different monitoring algorithms under two scenarios are compared in Figure \ref{fig:fig2} a) and b). LinUCB algorithm exhibits the highest cumulative regret in both scenarios since it does not exploit the unit dependency. GOB.Lin algorithm achieves lower cumulative regret compared to LinUCB by leveraging the similarities between units. The proposed CL-UCB algorithms outperform existing multi-armed bandit algorithms in both high-constraint and relaxed-constraint scenarios by explicitly capturing the latent group structure embedded in the population. By integrating the latent group structure and similarities between units, the CL-UCB algorithm recommends better monitoring strategy compared to the one without similarity regularization. The difference is more significant in the relaxed-constraint scenario. This indicates that incorporating latent group structure and similarity information into the model yields the greatest performance improvement. The model learning accuracy (L2-norm difference) of different algorithms is summarized in Figure \ref{fig:fig2}c) and d). By exploiting the latent group structure and similarity between units, the proposed algorithms achieve more accurate estimation of model parameters and faster convergence compared to the existing multi-armed bandit algorithms.

To explore the sensitivity of model performance with respect to the population size ($N$) and the number of latent groups ($K$), we further conduct sensitivity analysis on these parameters in Figure \ref{fig:box1} and Figure \ref{fig:box2}. Figure \ref{fig:box1} illustrates the cumulative regret of monitoring strategies recommended from four algorithms under different population sizes. It shows that the cumulative regrets of LinUCB and GOB.Lin algorithms increase linearly when the population size increases. GOB.Lin attains some regret reduction compared with LinUCB because of the effects of similarity regularization. The cumulative regrets of two proposed algorithms (CL-UCB and CL-UCB without similarity regularization) are lower than the benchmark MAB algorithms and do not increase linearly as the population size increases. This indicates that the proposed methods are robust in both small and large populations by explicitly capturing the latent group structure and similarities between units. Figure \ref{fig:box2} illustrates the cumulative regret under different numbers of latent groups. When the number of latent groups increases ($K=5$), the cumulative regrets of all models increase significantly, which indicates that the online modeling and monitoring problem is more challenging under complex latent group structure. The difference between the proposed methods (CL-UCB and CL-UCB without similarity regularization) and benchmark MAB algorithms is more obvious under the complex latent group structure, which suggests that the proposed methods can achieve more stable performance than the benchmark models when the latent structure becomes complex. 

\subsection{Application to cognitive degradation monitoring of Alzheimer's Disease (AD)}
The effectiveness of the proposed algorithm is further demonstrated through 100 real-world patients acquired from Alzheimer's Disease Neuroimaging Initiative (ADNI) \citep{MUELLER2005869}. The dataset consists of longitudinal measurements of MMSE for 648 units. They are collected at baseline, $12^{th}, 24^{th}, 36^{th}, 48^{th}$, and $60^{th}$ month. The missing MMSE values for each unit are imputed using linear interpolation. The MMSE score is based on the severity of degradation which is classified into three states including healthy state (27-30), mild cognitive impairment (24-26), and dementia (0-23) \citep{mitchell2009meta}. In addition to MMSE measurements, baseline measurements of risk factors such as ApoE genotypes and regional brain volume measurements extracted from MRI \citep{MUELLER2005869} are used to calculate the similarity between subjects, using the heat kernel method. The measurement collected from baseline to $60^{th}$ month period is equally divided into 30000 cycles. 
\begin{figure}[!t]
\centering
\includegraphics[width=\columnwidth]{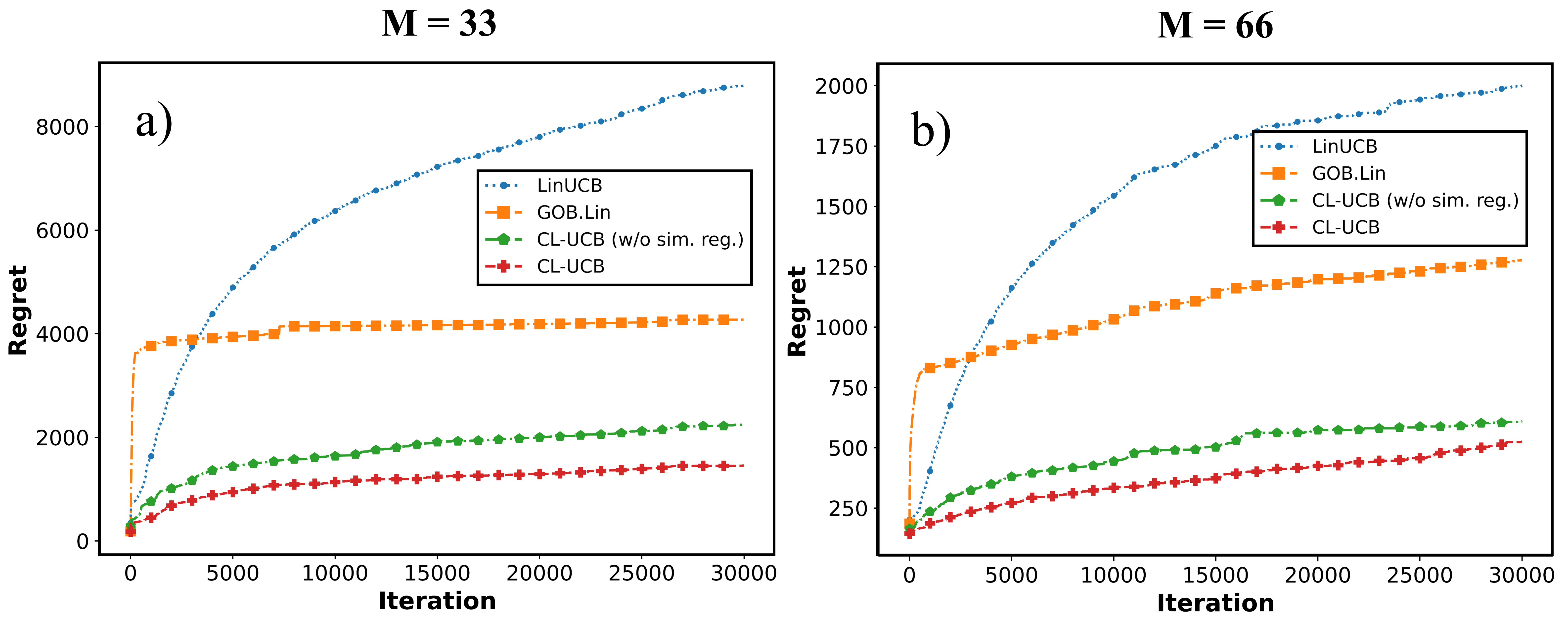}
\caption{The cumulative regret for all bandit algorithms shown in the Alzheimer’s Disease dataset with monitoring capacity of a) $M = 33$ and b) $M = 66$}
\label{fig:fig3}
\end{figure}

The cumulative regrets of different monitoring algorithms under two scenarios are compared in Figure \ref{fig:fig3}. It can be observed that LinUCB and GOB.Lin exhibit higher cumulative regret than the proposed algorithms in high-constraint scenario, suggesting that leveraging the latent structure of monitoring units effectively improves the accuracy of monitoring AD patients. CL-UCB has lower regret than CL-UCB without similarity regularization, indicating that incorporating both latent structure and similarity information into the model yields the best monitoring strategy for AD patients. Furthermore, the observations are consistent in the relaxed-constraint scenario which informs the efficiency of the proposed algorithms in monitoring both mild cognitive impairment and AD patients.

\section{Conclusion}
This paper introduces an Online Collaborative Learning (OCL) framework to adaptively allocate limited resources by modeling the dynamic health progression of dependent units and monitoring the high-risk ones. We develop a novel collaborative learning-based upper confidence bound (CL-UCB) algorithm to find an optimal monitoring strategy that balances the rewards of monitoring high-risk units (exploitation) with the uncertainty of the predicted health condition of unknown units (exploration). We present a regret analysis to prove a reduced upper regret bound compared to other benchmark models. Simulation studies and an empirical study of adaptive cognitive monitoring in Alzheimer’s disease demonstrate the efficiency of our proposed method as it achieves the lowest cumulative regret and highest learning accuracy compared to other benchmark models.

In the future, we will extend the proposed OCL framework to online model and monitor a population of dependent nonlinear health progression dynamics, such as kernel regressions. Novel UCB-based algorithm can be developed to balance the exploitation of high-risk processes and exploration of latent structure in kernel regressions. Moreover, the proposed OCL framework assumes the monitoring capacity is fixed in each monitoring cycle. In the future, we will relax this assumption by dynamically allocating the total monitoring budget to each monitoring cycle.


%
%
%
\newpage
\bibliographystyle{informs2014} 
\bibliography{References.bib} 

\newpage
\begin{APPENDICES}
\section{Proof of the Alternating Least Square Algorithm}
\begin{proof}{Proof.}
For $q$, let $L_1 = \sum_{t^{\prime}=1}^{t} \left\| C_{t}^{(d)T}X_{t^{\prime}}^Tq - y_{t^{\prime}} \right\|^{2} + \eta_1 \Vert q\Vert_2^2$, the partial derivative of $L_1$ with respect to $q$ is 
\[\frac{\partial L_1}{\partial q} = 2 \sum_{t^{\prime} = 1}^{t} X_{t^{\prime}}C_{t}^{(d)}(C_{t}^{(d)T}X_{t^{\prime}}^Tq - y_{t^{\prime}}) + 2\eta_1 q\]

Setting the left-hand side to zero, the equation above is arranged to be
\begin{align*}
    0 &= \sum_{t^{\prime} = 1}^{t} X_{t^{\prime}}C_{t}^{(d)}(C_{t}^{(d)T}X_{t^{\prime}}^Tq - y_{t^{\prime}}) + \eta_1 q\\
    0 &= \sum_{t^{\prime} = 1}^{t} (X_{t^{\prime}}C_{t}^{(d)}C_{t}^{(d)T}X_{t^{\prime}}^T + \eta_1 I_{Kp})q - \sum_{t^{\prime} = 1}^{t}  X_{t^{\prime}}C_{t}^{(d)}y_{t^{\prime}}\\
    \sum_{t^{\prime} = 1}^{t}  X_{t^{\prime}}C_{t}^{(d)}y_{t^{\prime}} &=   \sum_{t = 1}^{t} (X_{t^{\prime}}C_{t}^{(d)}C_{t}^{(d)T}X_{t^{\prime}}^T  + \eta_1 I_{Kp})q
\end{align*}
where $I_{Kp}$ is the identity matrix with the dimension of $Kp \times Kp$. The closed-form estimation of $\hat{q_{t}}$ with respect to the objective function above can be obtained by $\hat{q_t} = A_t^{-1} b_t$ where the calculation of $A_t$ and $b_t$ are computed as:
\begin{align*}
    A_t ={}& \sum_{t^{\prime} = 1}^{t} X_{t^{\prime}}C_{t}^{(d)}C_{t}^{(d)T}X_{t^{\prime}}^T  + \eta_1 I_{Kp} \\
    b_t ={}& \sum_{t^{\prime} = 1}^{t} X_{t^{\prime}}C_{t}^{(d)}y_{t^{\prime}}
\end{align*}
For $\Tilde{c}$, let $L_2 = \sum_{t^{\prime}=1}^{t} \left\| \Tilde{X}_{t^{\prime}}^T\Tilde{Q}_{t}\Tilde{c} - y_{t^{\prime}} \right\|^{2} + \eta_2\Vert \Tilde{c} \Vert_2^2$, the partial derivative of $L_2$ with respect to $\Tilde{c}$ is
\[\frac{\partial L_2}{\partial \Tilde{c}} = 2\sum_{t^{\prime}=1}^{t} \Tilde{Q}_{t}^T\Tilde{X_{t^{\prime}}}(\Tilde{X}_{t^{\prime}}^T\Tilde{Q}_{t}\Tilde{c} - y_{t^{\prime}}) + 2\eta_2\Tilde{c}\]

Setting the left-hand side to zero, the equation above is arranged to be
\begin{align*}
    0 &= \sum_{t^{\prime}=1}^{t} \Tilde{Q}_{t}^T\Tilde{X_{t^{\prime}}}(\Tilde{X}_{t^{\prime}}^T\Tilde{Q}_{t}\Tilde{c} - y_{t^{\prime}}) + \eta_2\Tilde{c}\\
    \sum_{t^{\prime}=1}^{t} \Tilde{Q}_{t}^T\Tilde{X_{t^{\prime}}} y_{t^{\prime}} &= (\sum_{t^{\prime}=1}^{t} \Tilde{Q}_{t}^T\Tilde{X_{t^{\prime}}}\Tilde{X}_{t^{\prime}}^T\Tilde{Q}_{t} + \eta_2I_{NK})\Tilde{c}
\end{align*}
where $I_{NK}$ is the identity matrix with the dimension of $NK \times NK$. The closed-form estimation of $\Tilde{c}$ with respect to the objective function above can be obtained by $\hat{\Tilde{c}}_t = D_t^{-1} d_t$ where the calculation of $D_t$ and $d_t$ are computed as:
\begin{align*}
    D_t ={}& \sum_{t^{\prime}=1}^{t} \Tilde{Q}_{t}^T\Tilde{X_{t^{\prime}}}\Tilde{X_{t^{\prime}}^T}\Tilde{Q}_{t} + \eta_2I_{NK} \\
    d_t ={}& \sum_{t^{\prime}=1}^{t} \Tilde{Q}_{t}^T\Tilde{X_{t^{\prime}}} y_{t^{\prime}}
\end{align*}
\end{proof}

\section{Proof of Lemma \ref{lemma1}}
\begin{proof}{Proof.}
For $q$, when $\Vert q_t\Vert_2 \leq L$ and $\Vert X_tC_t^{(d)}\Vert_2 \leq \Vert X_t\Vert_2 \Vert C_t^{(d)}\Vert_2 \leq SP$,
\begin{align*}
    A_t(\hat{q_t} - q^*) ={} A_t(A_t^{-1}b_t) - A_tq^*  ={} b_t - A_tq^* &={} \sum_{t^{\prime} = 1}^{t} X_{t^{\prime}}\hat{C}_{t^{\prime}}^{(d)}y_{t^{\prime}} -(\sum_{t^{\prime} = 1}^{t} X_{t^{\prime}}\hat{C}_{t^{\prime}}^{(d)}\hat{C}_{t^{\prime}}^{(d)T}X_{t^{\prime}}^T  + \eta_1 I_{kp})q^*\\
    &={} \sum_{t^{\prime} = 1}^{t} X_{t^{\prime}}\hat{C}_{t^{\prime}}^{(d)}\hat{C}_{t^{\prime}}^{(d)T}q^* + \epsilon) -\sum_{t^{\prime} = 1}^{t} X_{t^{\prime}}\hat{C}_{t^{\prime}}^{(d)}\hat{C}_{t^{\prime}}^{(d)T}X_{t^{\prime}}^T q^* -\eta_1 q^* \\
    &={} \sum_{t^{\prime} = 1}^{t} X_{t^{\prime}}\hat{C}_{t^{\prime}}^{(d)} \epsilon  + \sum_{t^{\prime} = 1}^{t} X_{t^{\prime}}\hat{C}_{t^{\prime}}^{(d)}(C^{(d)*T} - \hat{C}_{t^{\prime}}^{(d)T})X_{t^{\prime}}^Tq^* - \eta_1 q^*
\end{align*}
in which $\epsilon$ is the Gaussian noise in reward generation. The function norm of $\hat{q_t} - q^*$ is bounded by
\begin{align*}
    \Vert\hat{q_t} - q^*\Vert_{A_t} &={} \Vert \sum_{t^{\prime} = 1}^{t} X_{t^{\prime}}\hat{C}_{t^{\prime}}^{(d)} \epsilon  + \sum_{t^{\prime} = 1}^{t} X_{t^{\prime}}\hat{C}_{t^{\prime}}^{(d)}(C^{(d)*T} - \hat{C}_{t^{\prime}}^{(d)T})X_{t^{\prime}}^Tq^* - \eta_1 q^* \Vert_{A_t^{-1}} \\
    &\leq{} \Vert \sum_{t^{\prime} = 1}^{t} X_{t^{\prime}}\hat{C}_{t^{\prime}}^{(d)} \epsilon \Vert_{A_t^{-1}}  + \Vert \sum_{t^{\prime} = 1}^{t} X_{t^{\prime}}\hat{C}_{t^{\prime}}^{(d)}(C^{(d)*T} - \hat{C}_{t^{\prime}}^{(d)T})X_{t^{\prime}}^Tq^*  \Vert_{A_t^{-1}}  + \Vert \eta_1 q^* \Vert_{A_t^{-1}}\\
    &\leq{} \Vert \sum_{t^{\prime} = 1}^{t} X_{t^{\prime}}\hat{C}_{t^{\prime}}^{(d)}  \epsilon \Vert_{A_t^{-1}} + \frac{SPL}{\sqrt{\eta_1}}\sum_{t^{\prime} = 1}^{t}\Vert C^{(d)*T} - \hat{C}_{t^{\prime}}^{(d)T}\Vert_2 + \sqrt{\eta_1}L
\end{align*}
where the first term on the right-hand side of the inequality is bounded by the property of Theorem 1 in \citep{NIPS2011_e1d5be1c}. The second term follows the $q$-linear convergence as in \citep{10.1145/2983323.2983847}. For every $\epsilon_1 > 0$, $\Vert \hat{C}_{t+1}^{(d)T} - C^{(d)*}\Vert_2 \leq (v_2 + \epsilon_2)\Vert \hat{C}_{t}^{(d)T} - C^{(d)*}\Vert_2 $ where $0 < v_1 < 1$. Thus, for any $\delta > 0$, with probability at least $1 - \delta$, 
\begin{equation}
    \Vert\hat{q_t} - q^*\Vert_{A_t} \leq \sqrt{2\ln(\frac{\det(A_t)^{1/2}\det(\eta_1I)^{-1}}{\delta})} 
    + \frac{2SPL}{\sqrt{\eta_1}}\frac{(v_1+\epsilon_1)(1-(v_1+\epsilon_1)^t)}{1-(v_1+\epsilon_1)} + \sqrt{\eta_1}L \notag
\end{equation}

Since $\operatorname{Tr}(A_t) \leq \eta_1Kp + \sum_{t^{\prime} = 1}^{t}\operatorname{Tr}(X_{t^{\prime}}\hat{C}_{t^{\prime}}^{(d)}(X_{t^{\prime}}\hat{C}_{t^{\prime}}^{(d)})^T) \leq \eta_1Kp + \sum_{t^{\prime} = 1}^{t}\Vert X_{t^{\prime}}\Vert_2^{2} \Vert C_{t^{\prime}}^{(d)}\Vert_2^{2} \leq \eta_1Kp + tS^2P^2$, then $\det(A_t) \leq (\frac{\operatorname{Tr}(A_t)}{Kp})^{Kp} \leq (\eta_1 + \frac{tS^2P^2}{Kp})^{Kp}$ and $\det(\eta_1I) \leq \eta_1^{Kp}$. Putting all terms into the equation above, we have
\begin{equation}
    \Vert\hat{q_t} - q^*\Vert_{A_t} \leq \sqrt{Kp\ln(\frac{\eta_1 Kp + tS^2P^2}{\eta_1 Kp\delta})}  
    + \frac{2SPL}{\sqrt{\eta_1}}\frac{(v_1+\epsilon_1)(1-(v_1+\epsilon_1)^t)}{1-(v_1+\epsilon_1)} + \sqrt{\eta_1}L \notag
\end{equation}
For $\Tilde{c}$, 
\begin{align*}
    D_t(\hat{\Tilde{c_t}} - \Tilde{c^*}) ={} D_t(D_t^{-1}d_t) - D_t\Tilde{c^*}
    ={} d_t - D_t\Tilde{c^*}
    ={}& \sum_{t^{\prime}=1}^{t} \hat{\Tilde{Q}}_{t^{\prime}}^T\Tilde{X_{t^{\prime}}} y_{t^{\prime}} - (\sum_{t=1^{\prime}}^{t} \hat{\Tilde{Q}}_{t^{\prime}}^T\Tilde{X_{t^{\prime}}}\Tilde{X^T_{t^{\prime}}}\hat{\Tilde{Q}}_{t^{\prime}} + \eta_2 I_{NK})\Tilde{c^*}\\
    ={}& \sum_{t^{\prime}=1}^{t} \hat{\Tilde{Q}}_{t^{\prime}}^T\Tilde{X_{t^{\prime}}} (\Tilde{X^T_{t^{\prime}}}\Tilde{Q}^*\Tilde{c}^* + \epsilon) - (\sum_{t=1^{\prime}}^{t} \hat{\Tilde{Q}}_{t^{\prime}}^T\Tilde{X_{t^{\prime}}}\Tilde{X^T_{t^{\prime}}}\hat{\Tilde{Q}}_{t^{\prime}} + \eta_2 I_{NK})\Tilde{c^*}\\
    ={}& \sum_{t^{\prime}=1}^{t} \hat{\Tilde{Q}}_{t^{\prime}}^T\Tilde{X_{t^{\prime}}}\epsilon + \sum_{t^{\prime}=1}^{t} \hat{\Tilde{Q}}_{t^{\prime}}^T\Tilde{X_{t^{\prime}}}\Tilde{X^T_{t^{\prime}}}(\Tilde{Q}^* - \hat{\Tilde{Q}}_{t^{\prime}})\Tilde{c^*} - \eta_2 \Tilde{c^*}   
\end{align*}
in which $\epsilon$ is the Gaussian noise in reward generation. The function norm of $\hat{\Tilde{c}}_t - \Tilde{c^*}$ is bounded by
\begin{align*}
        \Vert\hat{\Tilde{c_t}} - \Tilde{c^*}\Vert_{D_t} 
        &={} \Vert \sum_{t^{\prime}=1}^{t} \hat{\Tilde{Q}}_{t^{\prime}}^T\Tilde{X_{t^{\prime}}}\epsilon + \sum_{t^{\prime}=1}^{t} \hat{\Tilde{Q}}_{t^{\prime}}^T\Tilde{X_{t^{\prime}}}\Tilde{X^T_{t^{\prime}}}(\Tilde{Q}^* - \hat{\Tilde{Q}}_{t^{\prime}})\Tilde{c^*} - \eta_2 \Tilde{c^*}  \Vert_{D_t^{-1}} \\
        &\leq{} \Vert\sum_{t^{\prime}=1}^{t} \hat{\Tilde{Q}}_{t^{\prime}}^T\Tilde{X_{t^{\prime}}}\epsilon\Vert_{D_t^{-1}} + \Vert \sum_{t^{\prime}=1}^{t} \hat{\Tilde{Q}}_{t^{\prime}}^T\Tilde{X_{t^{\prime}}}\Tilde{X^T_{t^{\prime}}}(\Tilde{Q}^* - \hat{\Tilde{Q}}_{t^{\prime}})\Tilde{c^*} \Vert_{D_t^{-1}} + \Vert \eta_2 \Tilde{c^*}\Vert_{D_t^{-1}} \\
        &\leq{} \Vert \sum_{t^{\prime}=1}^{t}  \hat{\Tilde{Q}}_{t^{\prime}}^T\Tilde{X_{t^{\prime}}}\epsilon\Vert_{D_t^{-1}} + \frac{SPL}{\sqrt{\eta_2}}\sum_{t^{\prime}=1}^{t} \Vert\Tilde{Q}^* - \hat{\Tilde{Q}}_{t^{\prime}}\Vert_2 + \Vert \eta_2 \Tilde{c^*}\Vert_{D_t^{-1}}
\end{align*}
where the first term on the right-hand side of the inequality is bounded by the property of Theorem 1 in \citep{NIPS2011_e1d5be1c}. The second term follows the $q$-linear convergence as in \citep{10.1145/2983323.2983847}. For every $\epsilon_1 > 0$, $\Vert\hat{\Tilde{Q}}_{t+1} - \Tilde{Q}^*\Vert_2 \leq (v_1 + \epsilon_1)\Vert\hat{\Tilde{Q}}_t - \Tilde{Q}^*\Vert_2 $ where $0 < v_2 < 1$. The third term is based on $\Vert \Tilde{c} \Vert = \Tilde{c}^T F \Tilde{c} = L(c_1,\ldots,c_n)$. Thus, for any $\delta > 0$, with probability at least $1 - \delta$,
\begin{equation}
    \Vert\hat{\Tilde{c_t}} - \Tilde{c^*}\Vert_{D_t} \leq{} \sqrt{2\ln(\frac{\det(D_t)^{1/2}\det(\eta_1I)^{-1}}{\delta})} 
    + \frac{2SPL}{\sqrt{\eta_2}}\frac{(v_2+\epsilon_2)(1-(v_2+\epsilon_2)^t)}{1-(v_2+\epsilon_2)} + \eta_2 L(c_1,\ldots,c_n) \notag
\end{equation}


Since $\operatorname{Tr}(D_t) \leq \eta_2NK + \sum_{t^{\prime} = 1}^{t}\operatorname{Tr}(\hat{\Tilde{Q}}_{t^{\prime}}^T\Tilde{X_{t^{\prime}}}\Tilde{X^T_{t^{\prime}}}\hat{\Tilde{Q}}_{t^{\prime}}) \leq \eta_2NK + \sum_{t^{\prime} = 1}^{t}\operatorname{Tr} (F_\otimes^{-1}) \Vert q_{t^{\prime}}\Vert_2^{2} \Vert X_{t^{\prime}}\Vert_2^{2} \leq \eta_2NK + S^2L^2 \sum_{t^{\prime} = 1}^{t} \sum_{j = 1}^{NK} (f^{-1}_\otimes)_{j,j}$ where $(f^{-1}_\otimes)_{j,j}$ is the $(j, j)^{th}$ element of the inverse of $F_\otimes$, then we have $\det(D_t) \leq ( \frac{\operatorname{Tr}(D_t)}{NK})^{NK} \leq (\eta_2 + \frac{S^2L^2 \sum_{t^{\prime} = 1}^{t} \sum_{j = 1}^{NK} (f^{-1}_\otimes)_{j,j}}{NK})^{NK}$ and $\det(\eta_2I) \leq \eta_2^{NK}$. Putting all terms into the equation above, we have
\begin{equation}
    \Vert\hat{\Tilde{C_t}} - \Tilde{C^*}\Vert_{D_t} 
    \leq{} \sqrt{NK\ln(\frac{\eta_2NK + S^2L^2 \sum_{t^{\prime} = 1}^{t} \sum_{j = 1}^{NK} (f^{-1}_\otimes)_{j,j}}{\eta_2 NK\delta})} 
    + \frac{2SPL}{\sqrt{\eta_2}}\frac{(v_2+\epsilon_2)(1-(v_2+\epsilon_2)^t)}{1-(v_2+\epsilon_2)} + \eta_2 L(c_1,\ldots,c_n) \notag
\end{equation}

\end{proof}

\section{Proof of Theorem \ref{theorem1}}
\begin{proof}{Proof.}
Similar to $y_t$ in Equation \ref{eq:12} in Section \ref{sub3.4.2}, for each unit $i$, we define its health condition as $y_{t,i} = \Tilde{c}^{T}X_{t,i}^{T}q$. 
According to the regret ($R_{t,i}$) in Equation \ref{eq:13}, because $a_{t,i} = 1$ for the selected unit $i$ at time $t$, the regret for each unit $i$ can be written as, 
\begin{align*}
    R_{t,i} ={} r_{t,i}^* - r_{t,i} ={} y_{t,i}^* - y_{t,i} ={}& \Tilde{c}^{*T}X_{t,i^*}^{T}q^* - \Tilde{c}^{*T}X_{t,i}^{T}q^* \\
    \leq{}& \hat{\Tilde{c}}_t^{T}X_{t,i^*}^{T}\hat{q_{t}} + \alpha^{\Tilde{c}} \Vert X_{t,i^*}^{T}\hat{q_{t}} \Vert_{D_{t}^{-1}} 
    + \alpha^q \Vert \hat{\Tilde{c}}_t^{T}X_{t,i^*}^{T} \Vert_{A_{t}^{-1}} - \Tilde{c}^{*T}X_{t,i}^{T}q^* \\
    \leq{}& \hat{\Tilde{c}}_t^{T}X_{t,i}^{T}\hat{q_{t}} + \alpha^{\Tilde{c}} \Vert X_{t,i}^{T}\hat{q_{t}} \Vert_{D_{t}^{-1}} 
    + \alpha^q \Vert \hat{\Tilde{c}}_t^{T}X_{t,i}^{T} \Vert_{A_{t}^{-1}} - \Tilde{c}^{*T}X_{t,i}^{T}q^* \\
    \leq{}&  2\alpha^{\Tilde{c}} \Vert X_{t,i}^{T}\hat{q_{t}} \Vert_{D_{t}^{-1}} + 2\alpha^q \Vert \hat{\Tilde{c}}_t^{T}X_{t,i}^{T} \Vert_{A_{t}^{-1}} + 2v^{2t}
\end{align*}
The first inequality is due to the definition of the UCB algorithm and the second inequality is due to the UCB arm selection strategy. The third inequality is based on the following.
\begin{align*}
    \hat{\Tilde{c}}_t^{T}X_{t,i}^{T}\hat{q_{t}} - \Tilde{c}^{*T}X_{t,i}^{T}q^* 
    ={}&  \hat{\Tilde{c}}_t^{T}X_{t,i}^{T}\hat{q_{t}}  - \hat{\Tilde{c}}_t^{T}X_{t,i}^{T}q^* + \hat{\Tilde{c}}_t^{ T}X_{t,i}^{T}q^* - \Tilde{c}^{*T}X_{t,i}^{T}q^* \\
    ={}& \hat{\Tilde{c}}_t^{T}X_{t,i}^{T}(\hat{q_{t}} - q^*) + (\hat{\Tilde{c}}_t^T - \Tilde{c}^{*T})X_{t,i}^{T}q^*\\ 
    ={}& \hat{\Tilde{c}}_t^{T}X_{t,i}^{T}(\hat{q_{t}} - q^*) + (\hat{\Tilde{c}}_t^T - \Tilde{c}^{*T})X_{t,i}^{T}\hat{q_{t}} 
    - (\hat{\Tilde{c}}_t^T - \Tilde{c}^{*T})X_{t,i}^{T}\hat{q_{t}} + (\hat{\Tilde{c}}_t^T - \Tilde{c}^{*T})X_{t,i}^{T}q^*\\
    ={}& \hat{\Tilde{c}}_t^{T}X_{t,i}^{T}(\hat{q_{t}} - q^*) + (\hat{\Tilde{c}}_t^T - \Tilde{c}^{*T})X_{t,i}^{T}\hat{q_{t}} 
    + (\hat{\Tilde{c}}_t^T - \Tilde{c}^{*T})X_{t,i}^{T}(q^* - \hat{q_{t}}) 
\end{align*}
By Cauchy Schwarz’s Inequality,
\begin{align*}
   \leq{}& \Vert \hat{\Tilde{c}}_t^{T}X_{t,i}^{T} \Vert \Vert \hat{q_{t}} - q^* \Vert + \Vert \hat{\Tilde{c}}_t^T - \Tilde{c}^{*T} \Vert \Vert X_{t,i}^{T}\hat{q_{t}} \Vert 
   +{} \Vert \hat{\Tilde{c}}_t^T - \Tilde{c}^{*T} \Vert  \Vert X_{t,i}^{T} \Vert \Vert q^* - \hat{q_{t}} \Vert \\ 
    \leq{}& \Vert \hat{\Tilde{c}}_t^{T}X_{t,i}^{T} \Vert_{A_{t}^{-1}} \Vert \hat{q_{t}} - q^* \Vert_{A_{t}} + \Vert \hat{\Tilde{c}}_t^T - \Tilde{c}^{*}\Vert_{D_t} \Vert  X_{t,i}^{T}\hat{q_{t}} \Vert_{D_{t}^{-1}} 
    +{} \Vert \hat{\Tilde{c}}_t^T - \Tilde{c}^{*T} \Vert  \Vert X_{t,i}^{T} \Vert \Vert q^* - \hat{q_{t}} \Vert
\end{align*}
From the proof on \textbf{Lemma \ref{lemma1}}, $\Vert\hat{\Tilde{c}}_t - \Tilde{c^*}\Vert_{D_t} \leq \alpha^{\Tilde{c}}$ and $\Vert\hat{q_t} - q^*\Vert_{A_t} \leq \alpha^q$. Thus,
\begin{equation}
    \Tilde{c}^{*T}X_{t,i^*}^{T}q^* - \Tilde{c}^{*T}X_{t,i}^{T}q^*
    \leq{} 2\alpha^q \Vert \hat{\Tilde{c}}_t^{T}X_{t,i}^{T} \Vert_{A_{t}^{-1}} + 2\alpha^{\Tilde{c}} \Vert X_{t,i}^{T}\hat{q_{t}} \Vert_{D_{t}^{-1}} + 2v^{2t} \notag
\end{equation}
in which $v$ is a constant in the range of $(0, 1)$, and the third term is based on the $q$-linear convergence rate of parameter estimation and the proof of Lemma 3.3 in \cite{10.1145/3292500.3330874}. 

The accumulated regret ($R(T) = \sum_{t=1}^{T} \sum_{i \in S_t} R_{t, i}$) of the CL-UCB algorithm at time $T$ is derived as, 
\begin{align} 
    R(T) ={}& 2 \alpha^{\Tilde{c}} \sum_{t=1}^{T} \sum_{i \in S_t} \Vert  X_{t,i}^{T}\hat{q_{t}} \Vert_{D_{t}^{-1}} 
    +{} 2 \alpha^q \sum_{t=1}^{T} \sum_{i \in S_t}\Vert \hat{\Tilde{c}}_t^{T}X_{t,i}^{T} \Vert_{A_{t}^{-1}} + \sum_{t=1}^{T} \sum_{i \in S_t} 2v^{2t} \notag \\
    \leq{}& 2\sqrt{T (\alpha^{\Tilde{C}})^2 \sum_{t=1}^{T} \sum_{i \in S_t}\Vert X_{t,i}^{T}\hat{q_{t}} \Vert_{D_{t}^{-1}}^2} 
    +{} 2\sqrt{T (\alpha^q)^2 \sum_{t=1}^{T} \sum_{i \in S_t} \Vert \hat{\Tilde{c}}_t^{T}X_{t,i}^{T} \Vert_{A_{t}^{-1}}^2} + \sum_{t=1}^{T} \sum_{i \in S_t} 2v^{2t} \label{eq:14}
\end{align}
Based on Lemma 11 in \cite{NIPS2011_e1d5be1c} and \cite{qin2014contextual}, the first and second terms in the above inequality can be bounded by,
\begin{align}
    &2\sqrt{T (\alpha^{\Tilde{C}})^2 \sum_{t=1}^{T} \sum_{i \in S_t}\Vert X_{t,i}^{T}\hat{q_{t}} \Vert_{D_{t}^{-1}}^2} + 2\sqrt{T (\alpha^q)^2 \sum_{t=1}^{T} \sum_{i \in S_t}\Vert \hat{\Tilde{c}}_t^{T}X_{t,i}^{T} \Vert_{A_{t}^{-1}}^2} \notag\\
    &\leq 2\alpha^{\Tilde{c}}\sqrt{2T\ln{(1 + \frac{\det(D_t)}{\det(\eta_2I)})}} + 2\alpha^q\sqrt{2TKp\ln{(1 + \frac{\det(A_t)}{\det(\eta_1I)})}}\notag \\
    &\leq 2\alpha^{\Tilde{c}}\sqrt{2TNK\ln{(1 + \frac{S^2L^2 \sum_{t^{\prime} = 1}^{t} \sum_{j = 1}^{NK} (f^{-1}_\otimes)_{j,j}}{\eta_2 NK})}} + 2\alpha^q\sqrt{2TKp\ln{(1 + \frac{TS^2P^2}{\eta_1 Kp})}} \label{eq:15}
\end{align}

The third term is calculated to be
\begin{align} \label{eq:16}
     \sum_{t=1}^{T} \sum_{i \in S_t} 2q^{2t} = \frac{2mv^2(1-v^{2T})}{1-v^2} 
\end{align}
Putting Equation \ref{eq:16} and \ref{eq:15} together in Equation \ref{eq:14}, the regret bound of CL-UCB is bounded by,
\begin{equation}
    R(T) \leq 2\alpha^{\Tilde{c}}\sqrt{2TNK\ln{(1 + \frac{S^2L^2 \sum_{t^{\prime} = 1}^{t} \sum_{j = 1}^{NK} (f^{-1}_\otimes)_{j,j}}{\eta_2 NK})}} + 2\alpha^q\sqrt{2TKp\ln{(1 + \frac{TS^2P^2}{\eta_1 Kp})}} + \frac{2mv^2(1-v^{2T})}{1-v^2} \notag
\end{equation}

\end{proof}

\end{APPENDICES}




\end{document}